\documentclass[runningheads]{llncs}

\usepackage{eccv}

\usepackage{eccvabbrv}

\usepackage{graphicx}
\usepackage{booktabs}
\usepackage{wrapfig}
\usepackage[accsupp]{axessibility}  

\usepackage{amsmath}
\usepackage{amssymb}
\usepackage{mathtools}
\usepackage[algo2e,linesnumbered,ruled,vlined]{algorithm2e}

\usepackage{graphicx}%
\usepackage{multirow}%
\usepackage{amsmath,amssymb,amsfonts}%
\usepackage{mathrsfs}%
\usepackage[title]{appendix}%
\usepackage{xcolor}%
\usepackage{textcomp}%
\usepackage{manyfoot}%
\usepackage{booktabs}%
\usepackage{listings}%
\usepackage{bm}%
\usepackage{multirow}%
\usepackage{color}
\usepackage{colortbl}
\usepackage{tabularx}
\usepackage[table]{xcolor}

\usepackage{booktabs}
\usepackage{multirow}
\usepackage{tabularx}
\usepackage{makecell}
\usepackage{amssymb}
\usepackage{pifont}
\usepackage{amsmath}
\usepackage{tcolorbox}
\usepackage{listings}
\usepackage{xcolor}
\tcbuselibrary{skins}
\usepackage{hyperref}
\usepackage{float}
\usepackage{tcolorbox}
\tcbuselibrary{breakable}
\usepackage[utf8]{inputenc} 
\usepackage[T1]{fontenc}    
\usepackage{url}            
\usepackage{booktabs}       
\usepackage{amsfonts}       
\usepackage{nicefrac}       
\usepackage{microtype}      
\usepackage{array}
\newcolumntype{C}[1]{>{\centering\arraybackslash}m{#1}}
\usepackage{xcolor}         
\usepackage{color, colortbl}
\definecolor{citecolor}{HTML}{2980b9}
\definecolor{linkcolor}{HTML}{c0392b}
\definecolor{darkorange}{HTML}{FF8C00}
\definecolor{chocolate}{HTML}{D2691E}
\definecolor{darkgreen}{HTML}{006400}
\definecolor{darkblue}{HTML}{00008B}
\definecolor{mediumblue}{HTML}{0000CD}
\definecolor{dodgerblue}{HTML}{1E90FF}
\definecolor{royalblue}{HTML}{4169E1}
\definecolor{shadecolor}{RGB}{237,237,237}
\definecolor{backred}{RGB}{255, 190, 190}
\definecolor{backblue}{RGB}{210, 230, 250}
\usepackage{colortbl}
\definecolor{graybg}{gray}{0.9}
\definecolor{zrrgreen}{HTML}{008000}
\definecolor{zrrblue}{HTML}{4682B4}
\definecolor{zrrred}{HTML}{B22222}
\definecolor{purple1}{RGB}{126, 107, 196}
\definecolor{purple2}{RGB}{199, 158, 207}
\definecolor{purple3}{RGB}{214, 200, 255}
\definecolor{purple4}{RGB}{254, 240, 255}
\usepackage{amsmath}
\usepackage{amssymb}
\usepackage{graphicx}

\usepackage{multirow}
\usepackage{makecell}
\usepackage{caption}

\usepackage{adjustbox}
\usepackage{wrapfig}

\usepackage{float}

\usepackage{hyperref}

\usepackage{orcidlink}

\begin{document}

\title{Code-in-the-Loop Forensics: Agentic Tool Use for Image Forgery Detection} 


\author{Fanrui Zhang\inst{1,2}\textsuperscript{$\dagger$} \orcidlink{0000-0002-1078-430X} \and
Qiang Zhang\inst{1}\textsuperscript{$\dagger$} \orcidlink{0000-0002-7809-5818} \and
Sizhuo Zhou\inst{1,2} \orcidlink{0009-0001-1147-4469} \and
Jianwen Sun\inst{2} \orcidlink{0009-0000-3145-5225} \and
Chuanhao Li\inst{3} \orcidlink{0000-0002-5769-3739} \and
Jiaxin Ai\inst{2} \orcidlink{0000-0002-4303-7749} \and
Yukang Feng\inst{2} \orcidlink{0009-0000-5594-3629} \and
Yujie Zhang\inst{2} \orcidlink{0009-0003-7069-0318} \and
Wenjie Li\inst{2} \orcidlink{0009-0000-4935-3382} \and
Zizhen Li\inst{2} \orcidlink{0009-0002-6554-9734} \and
Yifan Chang\inst{1,2} \orcidlink{0000-0003-3174-8540} \and
Jiawei Liu\inst{1}\textsuperscript{*} \orcidlink{0000-0001-9940-6366} \and
Kaipeng Zhang\inst{2,3}\textsuperscript{*} \orcidlink{0000-0001-6105-6532}
}

\authorrunning{F. Zhang et al.}

\institute{University of Science and Technology of China, China \and
Shanghai Innovation Institute, China \and
Shanghai Artificial Intelligence Laboratory, China\\
\email{zfr888@mail.ustc.edu.cn, jwliu6@ustc.edu.cn, kp\_zhang@foxmail.com}}

\begingroup
\renewcommand{\thefootnote}{$\dagger$}\footnotetext[1]{Equal contribution.}
\renewcommand{\thefootnote}{*}\footnotetext[1]{Corresponding author.}
\endgroup

\maketitle

\begin{abstract}
Existing image forgery detection (IFD) methods either exploit low-level, semantics-agnostic artifacts or rely on multimodal large language models (MLLMs) with high-level semantic knowledge.
Although naturally complementary, these two information streams are highly heterogeneous in both paradigm and reasoning, making it difficult for existing methods to unify them or effectively model their cross-level interactions.
To address this gap, we propose ForenAgent, a multi-round interactive IFD framework that enables MLLMs to autonomously invoke, execute, and iteratively refine Python-based low-level tools around the detection objective, thereby achieving more flexible and interpretable forgery analysis.
ForenAgent adopts a two-stage training pipeline with Cold Start and Reinforcement Fine-Tuning to progressively improve tool interaction and reasoning adaptability.
We design a human-inspired dynamic reasoning loop with global perception, local focusing, iterative probing, and holistic adjudication, and implement it as both a data-sampling strategy and a task-aligned process reward.
To support training and evaluation, we built FABench, a heterogeneous agent-forensics dataset with 100k images and about 200k agent-interaction question-answer pairs.
Experiments show that ForenAgent exhibits emergent tool-use competence and reflective reasoning on challenging IFD tasks when assisted by low-level tools, charting a promising route toward general-purpose IFD. The code is available at \url{https://github.com/zfr00/ForenAgent}.


\end{abstract}

\section{Introduction}
\label{sec:intro}



Advances in image editing and easy-to-use software have made low-cost manipulation and synthesis widely accessible. This growing democratization greatly boosts personal expression but also enables the malicious fabrication of multimedia content.~\cite{rao2022towards,shao2024detecting,zhu2023face,qi2022principled,qiao2024fully}. 
As a result, Image Forgery Detection (IFD) has become a critical research frontier, essential for mitigating the societal risks of large-scale visual manipulation and preserving information integrity.




Researchers have proposed a wide range of deep learning-based image forgery detection methods, achieving strong performance across various benchmarks.
\begin{wrapfigure}{r}{0.5\textwidth} 
    \centering
    \includegraphics[width=0.48\textwidth]{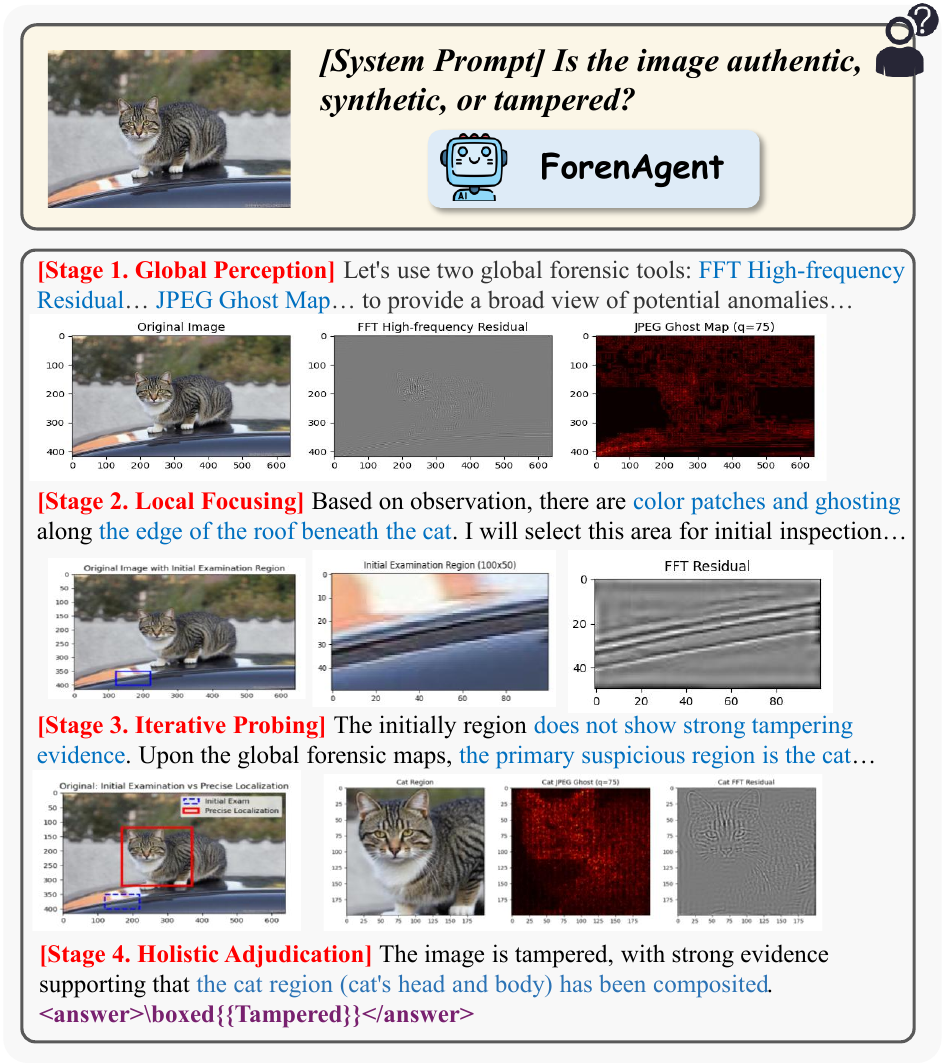}
    \caption{ForenAgent autonomously composes a global-to-local Python toolchain, accurately delivers a tampered verdict with precise localization of the forged region, and further demonstrates reflective self-correction by carefully revising an initially mislocalized crop to the appropriate region of interest.}
    \label{fig:fig1}
    \vspace{-5mm}
\end{wrapfigure}
Current approaches can be broadly categorized into two paradigms:
(1) Low-level feature-based methods: These approaches identify forgeries by capturing non-semantic inconsistencies between manipulated and authentic regions, focusing on subtle visual artifacts. Depending on the characteristics of the forged image, a wide range of low-level cues, such as JPEG compression artifacts, edge discontinuities, and camera model traces, have been utilized to enhance forensic perception. Such methods embody careful algorithmic design and strong domain priors, offering interpretability and effectiveness in specific scenarios.
However, depending solely on low-level inconsistencies restricts these methods to simple artifact patterns, making it difficult for them to handle diverse or subtle manipulation scenarios.
(2) MLLMs-based approaches: Recently, Multimodal Large Language Models (MLLMs) have achieved significant progress on tasks requiring integrated visual and textual understanding \cite{survey-mllm}. Methods such as FakeShield~\cite{xu2024fakeshield} and SIDA~\cite{huang2025sidasocialmediaimage} fine-tune MLLMs for IFD and demonstrate strong potential, benefiting from large-scale data to learn generalizable representations. 
Nonetheless, these approaches still exhibit several critical limitations: weak interaction with forensic tools, limited capability in fine-grained manipulation analysis, and insufficient transparency and controllability in sensitive scenarios. Fundamentally, these issues arise because their end-to-end learning paradigm does not encode structured forensic procedures or explicit tool-aware reasoning mechanisms.


Recent progress in MLLMs has shown that they are increasingly capable of complex reasoning and interaction with external tools~\cite{zhao2025pyvision,zheng2025deepeyes,jaech2024openai}. 
However, extending this mechanism to image forensics remains challenging: Current MLLMs lack a dynamic framework that connects high-level semantic reasoning with the control and interpretation of diverse low-level forensic tools, making task-adaptive integration difficult. Moreover, designing a training paradigm that guides the model toward logically consistent, self-directed reasoning and purposeful tool use rather than passive imitation remains an open challenge. Addressing these challenges is key to building truly interpretable and highly adaptive intelligent forensic systems.

In this work, we propose ForenAgent, an interactive multi-turn framework that enables MLLMs to autonomously invoke, execute, and iteratively refine Python-based low-level tools for IFD. To this end, we abstract and generalize common low-level forensics operators, including frequency residual, noise residual, and high-pass filtering, and package them into a toolbox of 12 utilities for future extension. 
For efficiency, only basic image processing operations, such as cropping, require code generation, while the remaining low-level forensics tools are exposed via direct tool calls.
As illustrated in Figure~\ref{fig:fig1}, ForenAgent autonomously orchestrates Python tools to verify a forged image from global screening to local inspection, ultimately classifying it as tampered and accurately localizing the forged region. The agent further demonstrates reflective self-correction by recovering from an initially misfocused crop to the correct area of interest, an “aha moment” observed in IFD agents.

The development of ForenAgent involves two key components:
(1) Forgery Agent Benchmark (FABench), a high-quality and heterogeneous forensic agent dataset constructed using state-of-the-art generative models (\textit{e.g.}, GPT-4o~\cite{achiam2023gpt}, Nano-Banana~\cite{nano}, and Midjourney-v7~\cite{midjourney2023})). It contains 100k images (40k real, 30k synthetic, and 30k tampered) and serves as a comprehensive benchmark for training and evaluation in IFD.
(2) A Cold-Start and Reinforcement Fine-Tuning (RFT) framework, designed to train MLLMs to function as reliable and autonomous agents. 
During the Cold-Start stage, ForenAgent adopts a self-exploration and experience-distillation paradigm. Specifically, GPT-4.1~\cite{openai2025gpt41} observes a large collection of forgery samples from FABench under system prompts that provide procedural guidance and executable code examples, distilling operational patterns into structured agent–interaction training data for initialization.
During RFT, we abstract the human IFD workflow into four reasoning stages: global perception, local focusing, iterative probing, and holistic adjudication. Correspondingly, we design four Forgery Process Rewards that together form the overall tool reward. By incorporating these verifiable reward components into the reinforcement learning process, ForenAgent develops a more interpretable and systematic forensic reasoning mechanism, effectively integrating basic image processing with low-level forensic analysis in a coherent investigative workflow.
This design enables ForenAgent to explore diverse reasoning strategies and optimize for long-term process quality rather than simply imitating predefined answers.

Extensive experiments demonstrate that ForenAgent significantly outperforms existing state-of-the-art IFD methods. Moreover, the model exhibits emergent multimodal reasoning behaviors such as visual search for forged regions, cross-region comparison, and even self-reflective correction. These intertwined reasoning patterns resemble human cognitive processes, contributing to stronger interpretability and forensic reliability for the IFD task.
Our main contributions are summarized as follows:

(1) We propose ForenAgent, a novel interactive, multi-turn framework that enables an MLLM to autonomously invoke, execute, and iteratively refine Python-based low-level tools for image forgery detection, thereby taking the first step toward intelligent, tool-augmented IFD systems.
(2) We construct FABench, a large-scale, high-quality, and heterogeneous forensic agent dataset comprising 100k images and 200k interactive QA pairs, focusing on forgery detection from cutting-edge generative models.
(3) We formulate a dynamic reasoning loop comprising global perception, local focusing, iterative probing, and holistic adjudication into a data-sampling strategy and task-aligned process reward that foster flexible, tool-adaptive evidence reasoning and enable robust, interpretable decisions.

\section{Related Work}
\label{sec:formatting}

\begin{figure*}[t]
    \centering
    \includegraphics[width=1\textwidth]{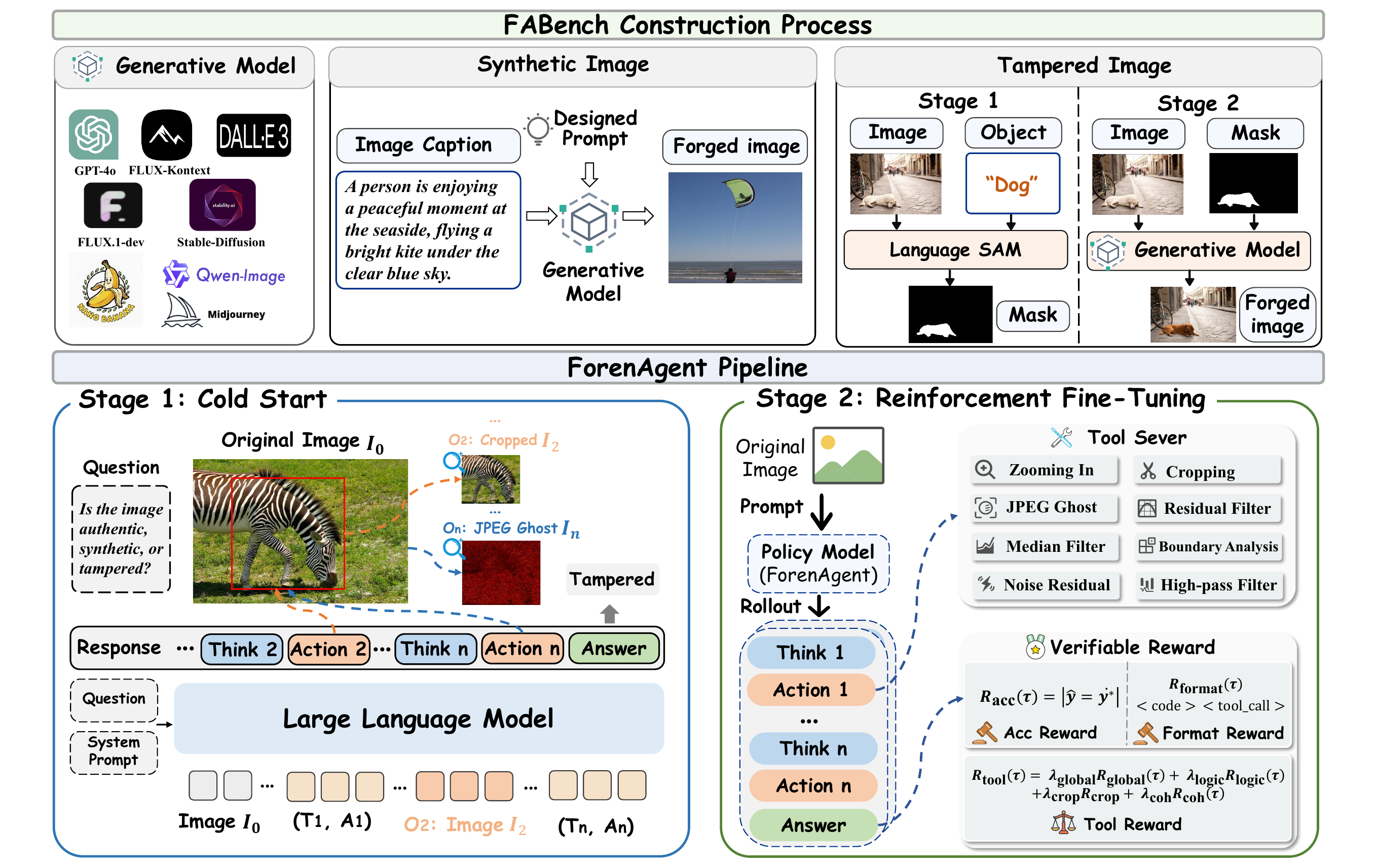}
    \caption{The overall architecture of the ForenAgent is illustrated, with the upper part showing the FABench construction process and the lower part presenting the training pipeline of ForenAgent.}
    \label{fig:overoframework}
    \vspace{-3mm}
\end{figure*}



\subsection{Image Forgery Detection}
Early IFD research primarily relies on low-level feature extractors that model non-semantic inconsistencies between manipulated and authentic regions. Typical frequency and residual cues include FFT-based high-frequency modeling in FreqNet~\cite{tan2024frequency}, SRM-filtered noise residuals in RGB-N~\cite{luo2021generalizing}, multi-scale high-frequency noise suppression in HFF~\cite{luo2021generalizing}, and transformation-driven feature learning in IT-Detector~\cite{li2025improving}. Related signals further exploit camera fingerprints (PRNU)~\cite{zhang2023prnu}, inconsistent compression (JPEG ghost)~\cite{popescu2005exposing}, DCT high-pass localization (ObjectFormer)~\cite{Objectformerwang2022objectformer}, and boundary-sensitive operators (MVSS-Net, HiFi-Net)~\cite{MVSS-Net2021image,HiFi_IFDL}. Additional work broadens these cues and improves robustness~\cite{chen2022self,bayar2016deep,liu2024forgery}. 
ST-Trace analyzes images through spatial-temporal evolution patterns~\cite{wang2025spatial}.
Despite progress, generalization remains challenging: Chameleon highlights failures on high-quality synthetic images~\cite{yan2024sanity}, while Effort~\cite{yan2024orthogonal} improves robustness by separating content from artifact features, better isolating forgery traces.
More recently, vision-language models and LLM-augmented frameworks have been introduced to incorporate semantic reasoning and natural-language explanations. DD-VQA~\cite{zhang2024common} enables fine-tuning BLIP on tampering data to improve both detection and explanations; FakeShield~\cite{xu2024fakeshield} and ForgeryGPT~\cite{liu2024forgerygpt} leverage LLMs for multimodal understanding and interactive, explainable analysis. For synthetic image detection, FakeScope~\cite{li2025fakescope} and LEGION~\cite{kang2025legion} further demonstrate the promise of MLLMs, while SIDA~\cite{huang2025sidasocialmediaimage} and So-Fake-R1~\cite{huang2025so} unify tampered and synthetic detection in a multi-class setting to probe MLLM capabilities. Overall, the field is moving from artifact-centric detection toward multimodal reasoning with improved interpretability.
AIGI-Holmes \cite{zhou2025aigi} employs MLLMs for explainable AI-generated image detection, unifying high-level reasoning with fine-grained artifact analysis to boost generalization. Loki \cite{ye2024loki} establishes a comprehensive benchmark for synthetic data detection, exposing the disparity between MLLMs' high-level reasoning and fine-grained artifact perception.
In line with this, our work also focuses on comprehensive detection across both tampered and synthetic images.

\subsection{Thinking with Images}
The “Thinking with Images” paradigm is pushing MLLMs beyond passive visual description toward interactive and iterative agents that can actively manipulate visual inputs and refine their reasoning processes \cite{zhu2024intelligent}. Early work typically relied on predefined CoT formats and static toolsets (\textit{e.g.}, VisProg \cite{gupta2023visual} and ViperGPT \cite{suris2023vipergpt}) to prompt models to invoke fixed tools for specific vision tasks. However, such designs often limit flexibility and generalization to unseen or complex scenarios. To address this limitation, METATOOL \cite{wang2024metatool} introduces meta-task augmentation to improve tool mastery and transferability. In contrast, PyVision \cite{zhao2025pyvision} enables dynamic Python code generation to flexibly invoke and compose complex tools for more versatile visual reasoning.
Pushing autonomy further, recent studies optimize tool use through reinforcement learning (RL) \cite{liu2025visual}. DeepEyes \cite{zheng2025deepeyes} performs end-to-end RL with tool-oriented data selection and reward design, while DeepEyesV2 \cite{hong2025deepeyesv2} advances multimodal reasoning toward a more agentic paradigm, achieving state-of-the-art results on perception- and search-intensive tasks in RealX-Bench. Thyme \cite{zhang2025thyme} shifts the focus from static image perception to comprehensive reasoning by integrating temporal and logical context. V-TOOLRL \cite{su2025openthinkimg} directly maximizes task success based on tool-interaction feedback, and ReVPT \cite{zhou2025reinforced} adopts a two-stage scheme with multiple visualization tools to enhance general multimodal capabilities. Collectively, these methods demonstrate a transition from fixed tool pipelines toward adaptive visual agents that learn more effective tool-use strategies through interaction and feedback.
Despite significant progress in numerous domains, exploration within the specialized field of image forgery detection remains nascent. To address this gap, we propose ForenAgent, a forensic agent based on a "Code-in-the-Loop" design. ForenAgent actively generates and iteratively optimizes low-level forensic analysis code, thereby achieving accurate detection of forged images.

\section{Method}
In this section, as shown in Figure~\ref{fig:overoframework}, we first introduce FABench, a comprehensive benchmark consisting of multi-type, high-difficulty forgery images. We then describe ForenAgent's two-stage training framework.

\begin{figure}[!t]
\centering
\begin{minipage}[t]{0.48\linewidth}
    \centering
    \includegraphics[width=\linewidth]{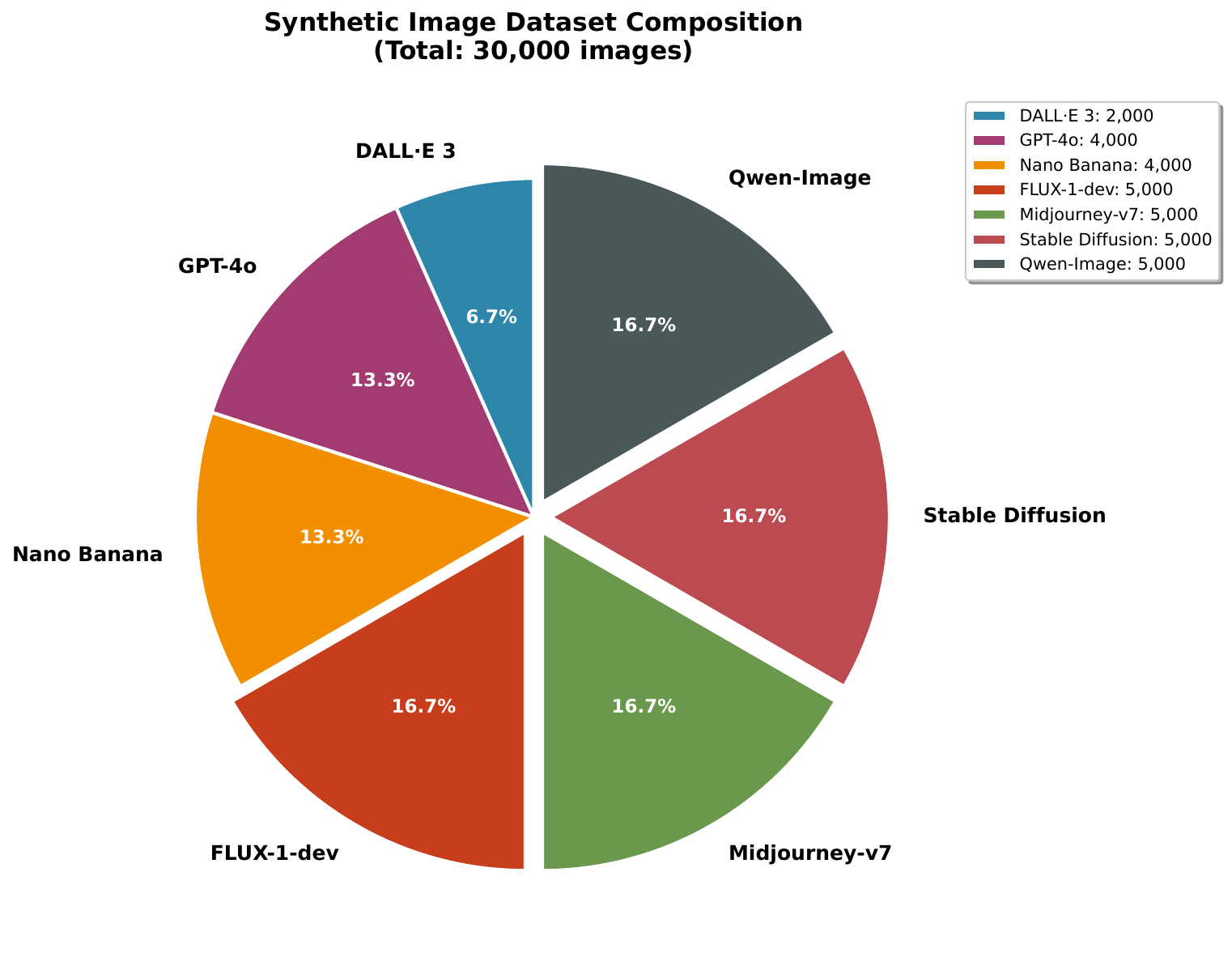}
\end{minipage}
\hfill
\begin{minipage}[t]{0.48\linewidth}
    \centering
    \includegraphics[width=\linewidth]{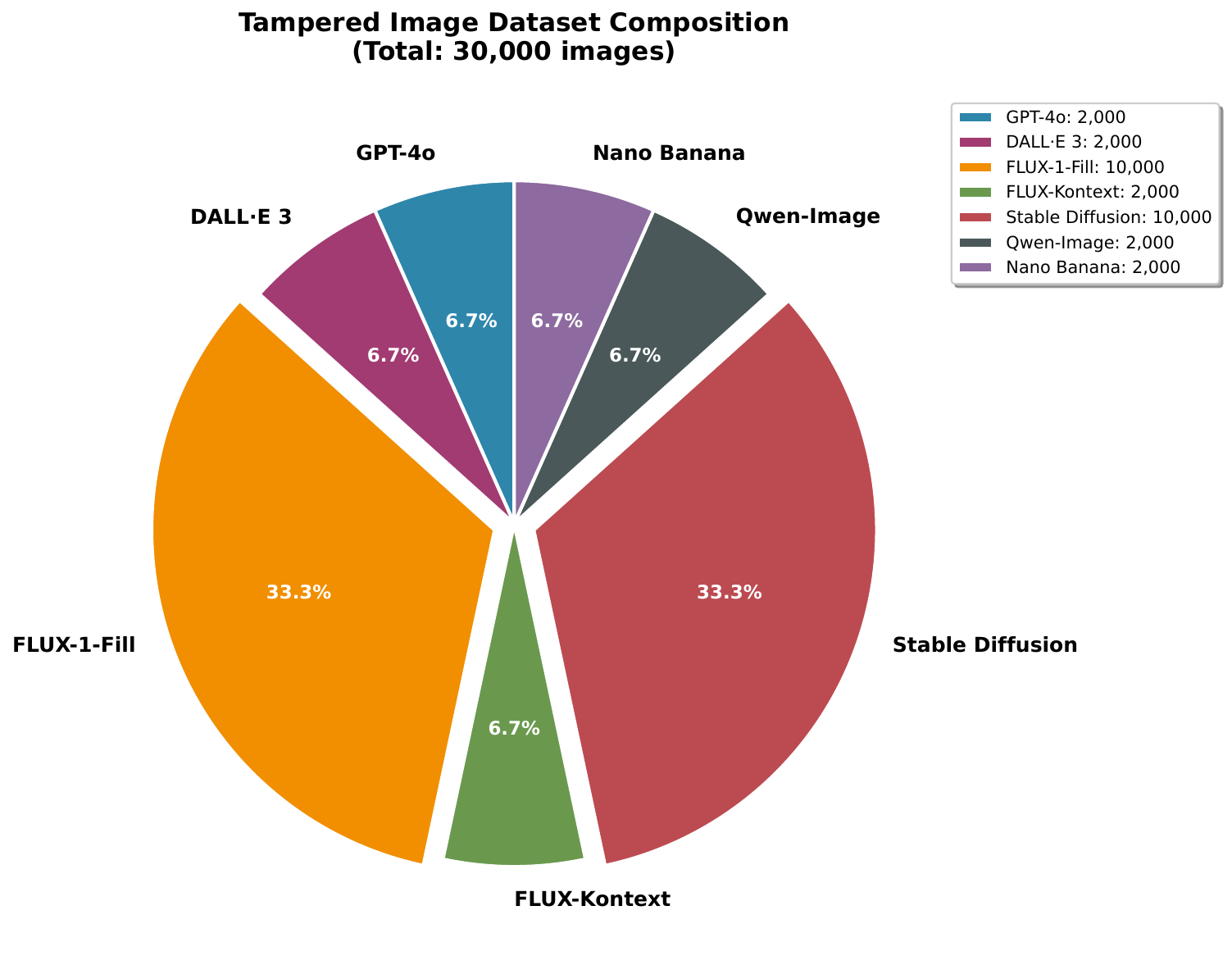}
\end{minipage}
\caption{The composition of the FABench training set.}
\label{fig:distribution_fabench}
\end{figure}

\begin{figure*}[t]
    \centering
    \includegraphics[width=1\textwidth]{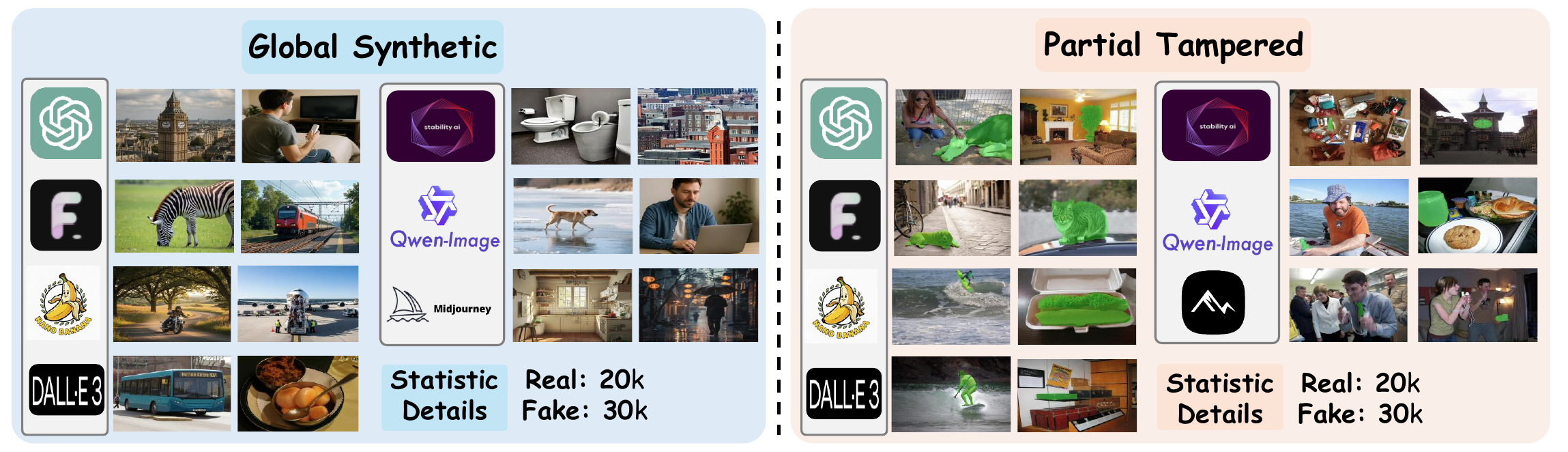}
    \caption{Examples of tampered and synthetic images from diverse FABench generators.}
    \label{fig:dataset}
    \vspace{-3mm}
\end{figure*}

\subsection{FABench}
Recent advances in generative AI have enabled the easy creation of sophisticated synthetic and tampered content, while existing detection datasets exhibit notable limitations:
(i) Outdated synthetic content. Benchmarks grounded in early GANs (\textit{e.g.}, StyleGAN~\cite{karras2019style}) mainly contain low-fidelity generations that are significantly easier than modern photorealistic outputs (\textit{e.g.}, GPT-4o-image~\cite{achiam2023gpt}, Midjourney-v7~\cite{midjourney2023}).
(ii) Fixed tampering pipelines. Many datasets rely on a narrow set of inpainting models (\textit{e.g.}, Stable Diffusion~\cite{sdxl}) and rarely explore newer pipelines (\textit{e.g.}, FLUX-Kontext~\cite{flux2024}, Qwen-image~\cite{wu2025qwenimage}), limiting heterogeneity and inducing repeated artifacts.
We build FABench via a strict, modular pipeline designed to maximize diversity across contemporary generators. FABench contains authentic, synthetic, and tampered images to reflect open-world scenarios and comprehensively evaluate forensic reasoning. Beyond simple objects and portraits, it also covers a wide range of naturalistic scenes:

{Authentic (40k)}: COCO~\cite{lin2014microsoft} images spanning a broad spectrum of real-world scenes and everyday contexts.

{Synthetic (30k)}: Two-step pipeline (as shown in Figure~\ref{fig:overoframework}): caption enrichment (30k COCO images; GPT-4o-mini generates detailed, compositional captions), followed by image synthesis with a diverse set of generators to maximize architectural diversity and realism.

{Tampered (30k)}: Starting from COCO sources with instance masks, we derive object masks via SAM-guided text prompts~\cite{lin2014microsoft}, then perform object-level inpainting using (i) strict-mask models (FLUX-1-Fill, Stable Diffusion; inputs: image/mask/prompt) and (ii) soft/no-mask models (\textit{e.g.}, GPT-4o-image, Qwen-Image), where the edited regions are composited back into the original image to suppress unintended global micro-changes. The comprehensive construction pipeline and detailed analysis are provided in the Appendix.

We adopt a multi-stage pipeline consisting of quality validation (file integrity, mask legality, etc.) and deduplication, followed by stratified human auditing to filter low-quality samples. Samples that fail inspection are removed, while borderline cases are re-synthesized or re-inpainted accordingly.
For the tampered split, we construct a 700-image tampered test set using seven generators: GPT-4o, DALL·E 3~\cite{openai2024dalle3}, FLUX-1-Fill, FLUX-Kontext, Stable Diffusion, Qwen-Image, and Nano Banana~\cite{nano}.
Similarly, for the synthetic split, we generate a 700-image synthetic test set using GPT-4o, DALL·E 3, FLUX-1-dev, Midjourney-v7, Stable Diffusion, Qwen-Image, and Nano Banana.
For the authentic split, we randomly sample 700 real images from COCO to form the authentic test set. 

We visualize the composition of the FABench training set in Figure~\ref{fig:distribution_fabench}.
For synthetic images, the dataset includes samples generated by GPT-4o (4 k), DALL·E 3 (2 k), FLUX-1-dev (5 k), Midjourney-v7 (5 k), Stable Diffusion (5 k), Qwen-Image (5 k), and Nano Banana (4 k).
For tampered images, it includes GPT-4o (2 k), DALL·E 3 (2 k), FLUX-1-Fill (10 k), FLUX-Kontext (2 k), Stable Diffusion (10 k), Qwen-Image (2 k), and Nano Banana (2 k).
This diverse composition demonstrates that FABench integrates a wide spectrum of generation and inpainting paradigms, covering both text-to-image and mask-based pipelines.
Figure~\ref{fig:dataset} further showcases examples of tampered and synthetic images produced by different generators in FABench. We observe that advanced models, such as Nano Banana and Qwen-Image, produce more photorealistic and harder-to-discriminate forgeries.


\subsection{ForenAgent}
ForenAgent enables MLLMs to dynamically invoke and execute low-level tools throughout their reasoning process. In each session, the MLLM receives the input and responds with either executable Python code or structured tool calls, which are run in an isolated Python sandbox. The resulting outputs, including text and or visualizations, are then fed back into the MLLM context, allowing iterative multi-turn refinement until the final prediction is produced.
\subsubsection{Tool Boxes}
ForenAgent adopts Python code and tool calls as the fundamental primitives for tool construction. Accordingly, we categorize the tools into two major types as follows:

\noindent\textbf{(1) Basic Image Processing:} These two categories of tools form the basis for visual manipulation and perception. They enable the agent to clean, align, and highlight image content to improve downstream reasoning.

{Cropping:} For high-resolution or cluttered inputs, the agent typically crops and zooms into regions of interest. By reasoning about the coordinates, it effectively performs soft object localization and forensic analysis, directing attention to the most informative regions.

{Enhancement:} In visually subtle domains like tampered imaging, the agent autonomously applies adaptive contrast adjustments and multi-scale enhancements to amplify latent structures and imperceptible forensic traces, making them significantly more prominent.

\noindent\textbf{(2) Low-Level Forensics Tools:} Based on the related work, we constructed a candidate pool of 12 low-level, code-based forensic methods. The agent can generate and deploy these tools as needed. We categorize them as follows:

{(a) Frequency Domain Analysis}: Tools that analyze artifacts in transformed domains.
        (1) {FFT High-Frequency Residual:} Emphasizes forgery boundaries and texture anomalies in the frequency domain.
        (2) {DWT High-Frequency Subbands:} Uses wavelet decomposition to reveal high-frequency differences from synthesis or upsampling.
        (3) {Resampling Periodicity:} Detects spectral peaks introduced by interpolation (scaling/rotation).
        (4) {DCT-based High-Pass Filter:} Extracts high-frequency components to highlight edges and tampering traces.
    
{(b) Noise \& Residual Analysis}: Tools that extract subtle noise patterns typically suppressed by image content.
        (5) {SRM:} Uses a bank of high-pass and directional filters to extract robust noise residuals.
        (6) {Bayar Constrained Convolution:} Employs a specific convolutional kernel to suppress image content and amplify manipulation traces.
        (7) {PRNU (Photo-Response Non-Uniformity):} Extracts the camera sensor's unique fingerprint noise to find local inconsistencies (splices) via block correlation.
    
{(c) Edge \& Boundary Analysis}: Methods to pinpoint inconsistent edges or gradients.
        (8) {Sobel Edge Detector:} Identifies splicing boundaries or anomalous edge patterns.
        (9) {Laplacian High-Pass:} Extracts high-frequency components to detect tampering artifacts.
    
{(d) Specific Artifact Detection}: Tools targeting the byproducts of distinct manipulations.
        (10) {JPEG Ghost:} Detects recompression artifacts by analyzing the error layer difference between multiple compression qualities.
        (11) {Median Filtering Traces:} measures artifacts and suspicious smoothing patterns.
    
{(e) Statistical Analysis}:
        (12) {Local Correlation Map:} Quantifies enhanced correlations within pixel neighborhoods, often indicative of manipulation.

For efficiency, only basic image processing operations such as cropping require Python code generation and execution, while the remaining low level forensics tools are exposed via direct tool calls.

\subsubsection{Cold Start}
The training process for ForenAgent consists of two sequential stages, designed to progressively equip the MLLM with the capabilities to handle complex image forgery detection tasks.
\\
\textbf{System Prompt Design}:
To steer the MLLM’s reasoning, code generation, and low-level tool usage, ForenAgent employs a carefully designed system prompt in addition to user queries. The prompt defines how to access inputs, invoke tools, structure code, and return results: 
(i) prefer executable code or tool calls over free-form text; 
(ii) preload images or frames as \texttt{image\_clue\_i} (with resolution metadata) to enable direct referencing for operations such as cropping; 
(iii) standardize outputs via \texttt{print(...)} for text and \texttt{plt.show()} for visualizations; 
(iv) wrap each code block with \texttt{<code>...</code>} for reliable parsing, and each tool invocation with \texttt{<tool\_call>...</tool\_call>} to enable structured execution; 
(v) place the final class token inside \texttt{<answer>...</answer>} for consistent evaluation. 
This design produces parsable and executable outputs with reduced runtime errors. The full system prompt is provided in the Appendix.
\\
\textbf{Correct Reasoning Trajectories}:
Built on FABench, we curate a long-horizon, multi-turn instruction-tuning set for IFD to inject domain reasoning and long-CoT skills into open-source MLLMs. For each sample, we provide the System Prompt, question, and images to GPT-4.1 to obtain an authenticity judgment and a multi-step chain. We retain a response only if: (1) the predicted label is correct; (2) It includes a code sandbox that enforces the enclosure of executable Python within \texttt{<code>...</code>} and tool invocations within \texttt{<tool\_call>...</tool\_call>}; (3) for tampered cases, the answer explicitly names the forged object. 
The filtered subset, containing approximately 200k agent–interaction question–answer pairs, is used for supervised Cold-Start tuning.

\subsubsection{Reinforcement Fine-Tuning}

In this section, we investigate how MLLMs acquire tool-calling and reasoning capabilities without the need for supervised labels, leveraging pure RL for continuous self-improvement. End-to-end, outcome-rewarded RL jointly optimizes textual CoT and action planning over full trajectories. The agent interacts for multiple turns until producing an answer or exhausting the tool-call budget. 
States interleave text tokens $X$ and image tokens $I$; all {observation} tokens are inputs only and do not contribute to the loss.
\\
\textbf{Group Relative Policy Optimization (GRPO)}:
With GRPO~\cite{guo2025deepseek}, we sample a candidate set for each input, compute relative rewards through within-group normalization, and update the policy without a separate critic network. By employing clipped importance weights alongside a KL divergence penalty relative to a reference model, we stabilize the training process while effectively distilling preference signals from both model-generated and human-annotated answers.
\\
\textbf{Reward Modeling}: 
In addition to the correctness reward $R_{\text{acc}}(\tau)$ and the format reward $R_{\text{format}}(\tau)$ that ensures the use of valid \texttt{<code>},\texttt{<tool\_call>} and \texttt{<answer>} tags, we introduce a tool usage reward $R_{\text{tool}}$ to evaluate how effectively the model applies external tools. The tools are categorized into {Basic Image Processing} $\mathcal{T}_{\text{basic}}$ and {Low-Level Forensics} $\mathcal{T}_{\text{low}}$. The reward $R_{\text{tool}}(\tau)$ integrates four components to assess the logical use of these tools.
\\
\textit{(i) Global Forensic Priority ($R_{\text{global}}$):} Encourages the model to first apply low-level forensic tools for global image analysis before using basic image processing for local operations. $T$ represents the total number of interaction turns. Let $t$ denote the index of the current reasoning step and $a_t$ represent the action. Define the first-use steps:
\begin{equation}
\begin{aligned}
t_{\text{low}} &= \min\{t : a_t \in \mathcal{T}_{\text{low}}\},\\
t_{\text{basic}} &= \min\{t : a_t \in \mathcal{T}_{\text{basic}}\}.
\end{aligned}
\end{equation}
The global priority reward is:
\begin{equation}
R_{\text{global}}(\tau) =
[t_{\text{low}} < t_{\text{basic}}] \cdot \gamma^{t_{\text{low}} - 1},
\quad \gamma \in (0, 1).
\end{equation}
\\
\textit{(ii) Tool Logic ($R_{\text{logic}}$):} This component motivates the model to optimize its behavior by rewarding syntactically correct and logically coherent tool invocations.
\\
\textit{(iii) Crop Sensitivity ($R_{\text{crop}}$):}
Reward a single occurrence of \texttt{Crop} with class-specific weights.
Define the indicator
\begin{equation}
\mathbb{I}_{\text{crop}} \;=\; \mathbb{1}\!\left\{\exists\, t\in\{1,\dots,T\}:~ a_t=\texttt{Crop}\right\},
\end{equation}
\begin{equation}
R_{\text{crop}}(\tau)=
\begin{cases}
b_{\text{tamper}}\, \mathbb{I}_{\text{crop}}, & \text{if } \hat{y}=\texttt{tampered},\\[2pt]
b_{\text{auth}}\, \mathbb{I}_{\text{crop}},   & \text{if } \hat{y}=\texttt{authentic},\\[2pt]
b_{\text{syn}}\, \mathbb{I}_{\text{crop}},    & \text{if } \hat{y}=\texttt{synthetic}.
\end{cases}
\end{equation}
\\
\textit{(iv) Reasoning Coherence ($R_{\text{coh}}$):}
Reward a “locate-then-investigate” pair once (at most): a low-level tool immediately after \texttt{Crop} that consumes its output.
Let
\begin{equation}
\begin{aligned}
R_{\text{coh}}
&= \mathbb{1}\!\Bigl\{\exists\, t\in\{1,\dots,T-1\}:\; a_t=\texttt{Crop},\\
&\qquad a_{t+1}\in\mathcal{T}_{\text{low}},\; \mathrm{chain}(a_t,a_{t+1})\Bigr\}.
\end{aligned}
\end{equation}
\\
The tool usage reward aggregates the four sub-rewards:
\begin{equation}
\begin{aligned}
R_{\text{tool}}(\tau) = \lambda_{\text{global}} R_{\text{global}}(\tau) + \lambda_{\text{logic}} R_{\text{logic}}(\tau)\\ + \lambda_{\text{crop}} R_{\text{crop}}(\tau) + \lambda_{\text{coh}} R_{\text{coh}}(\tau).
\end{aligned}
\end{equation}
\\
Finally, the overall reward function $R$ is defined as:
\begin{equation}
\begin{aligned}
R(\tau) = \lambda_{\text{acc}} \cdot R_{\text{acc}}(\tau) + \lambda_{\text{format}} \\
\cdot R_{\text{format}}(\tau) + \lambda_{\text{tool}} \cdot R_{\text{tool}}(\tau).
\end{aligned}
\end{equation}

By incorporating these verifiable reward components into the reinforcement learning process, ForenAgent achieves more interpretable and systematic reasoning for IFD, effectively learning to leverage both basic image processing and low-level forensic analysis tools in a coherent investigative workflow.
\begin{table*}[t]
    \caption{Comparison of ForenAgent with state-of-the-art methods on FABench, reporting per-class and overall performance across three categories.}
    \centering
    \resizebox{\textwidth}{!}{
    \begin{tabular}{lcccccccccccccccccccc}
        \toprule
        \multirow{2}{*}{\textbf{Methods}} 
        & \multicolumn{4}{c}{\textbf{Authentic}} 
        & \multicolumn{4}{c}{\textbf{Synthetic}} 
        & \multicolumn{4}{c}{\textbf{Tampered}} 
        & \multicolumn{2}{c}{\textbf{Overall}} \\
        \cmidrule(r){2-5}\cmidrule(r){6-9}\cmidrule(r){10-13}\cmidrule(r){14-15}
        & Acc & Prec & Rec & F1
        & Acc & Prec & Rec & F1
        & Acc & Prec & Rec & F1
        & Acc & F1 \\
        \midrule
        Gemini2.5-flash~\cite{team2023gemini}
        & 47.2 & 36.2 & 98.6 & 52.9
        & 75.2 & 88.2 & 38.5 & 53.6
        & 68.2 & 75.0 & 3.9 & 7.4
        & 45.3 & 38.0 \\
        Gemini2.5-Pro~\cite{team2023gemini}
        & 46.9 & 38.3 & 96.9 & 54.9
        & 72.6 & 83.2 & 22.0 & 34.8
        & 70.0 & 74.8 & 15.3 & 25.4
        & 44.7 & 38.4 \\
        GPT-4o~\cite{achiam2023gpt}
        & 46.6 & 38.1 & 96.3 & 65.8
        & 72.5 & 83.5 & 21.7 & 34.5
        & 69.8 & 71.8 & 15.3 & 25.4
        & 44.4 & 38.1 \\
        GPT-o3-mini~\cite{openai2025o3mini}
        & 46.6 & 38.1 & 96.0 & 54.5
        & 72.5 & 83.5 & 21.7 & 34.5
        & 69.9 & 72.4 & 15.7 & 25.8
        & 44.5 & 38.3 \\
        GPT-4.1~\cite{openai2025gpt41}
        & 54.0 & 41.1 & 95.9 & 57.5
        & 80.0 & 90.5 & 46.6 & 61.5
        & 68.8 & 65.3 & 13.0 & 21.6
        & 51.4 & 46.9 \\
        GPT-5~\cite{openai2025gpt5}
        & 46.5 & 38.1 & 96.3 & 54.6
        & 72.6 & 83.6 & 21.9 & 34.7
        & 69.8 & 73.1 & 15.1 & 25.1
        & 44.5 & 38.1 \\
        \hline
        InternVL3-78B~\cite{zhu2025internvl3}
        & 46.7 & 37.3 & 87.9 & 52.4
        & 66.0 & 46.0 & 12.1 & 19.2
        & 67.9 & 54.9 & 20.9 & 30.2
        & 40.3 & 33.9 \\
        Qwen2.5-VL-72B~\cite{Qwen2.5-VL}
        & 63.3 & 47.4 & 90.7 & 62.3
        & 70.3 & 56.4 & 48.3 & 52.0
        & 66.3 & 47.8 & 11.0 & 17.9
        & 50.0 & 44.1 \\
        QVQ-72B-preview~\cite{qvq-72b-preview}
        & 59.5 & 43.3 & 69.3 & 53.3
        & 66.4 & 49.4 & 38.1 & 43.1 
        & 65.2 & 46.6 & 29.3 & 36.0
        & 45.6 & 44.1 \\
        InternVL2.5-78B-MPO~\cite{wang2024mpo}
        & 60.1 & 45.0 & 88.4 & 59.6
        & 64.1 & 44.4 & 30.1 & 35.9
        & 62.0 & 30.2 & 10.7 & 15.8
        & 43.1 & 37.1 \\
        Qwen3-VL-30B~\cite{Qwen3-VL}
        & 55.8 & 42.4 & 90.6 & 57.7
        & 65.4 & 45.5 & 19.7 & 27.5
        & 71.7 & 67.7 & 29.0 & 40.6
        & 46.4 & 41.9 \\
        \hline
        GPT-4.1 (with Tools)~\cite{openai2025gpt41}
        & 77.8 & 65.5 & 70.4 & 67.9
        & 75.1 & 64.1 & 57.9 & 60.8
        & 76.4 & 64.3 & 65.7 & 65.0
        & 64.7 & 64.6 \\
        Qwen2.5-VL-72B (with Tools)~\cite{Qwen2.5-VL}
        & 74.2 & 61.5 & 60.1 & 60.8
        & 71.2 & 56.7 & 57.4 & 57.1
        & 69.5 & 54.2 & 54.7 & 54.4
        & 57.4 & 57.4 \\
        Gemini2.5-Pro (with Tools)~\cite{team2023gemini}
        & 71.3 & 57.0 & 57.1 & 57.1
        & 69.2 & 54.1 & 50.9 & 52.4
        & 68.3 & 52.3 & 55.3 & 53.8
        & 54.4 & 54.4 \\
        \hline
        Qwen2.5-VL-7B~\cite{Qwen2.5-VL}
        & 86.2 & 80.2 & 78.0 & 79.1
        & 86.1 & 79.5 & 78.4 & 78.9
        & 87.1 & 79.5 & 82.7 & 81.1
        & 79.7 & 79.9 \\
        Qwen3-VL-8B~\cite{Qwen3-VL}
        & 80.1 & 66.2 & 81.9 & 73.2
        & 86.9 & 88.7 & 69.6 & 78.0
        & 85.2 & 78.4 & 76.9 & 77.6
        & 76.1 & 76.3 \\
        Gram-Net~\cite{DBLP:conf/cvpr/LiuQT20}
        & 75.7 & 58.4 & 94.4 & 72.2
        & 74.6 & 67.2 & 46.3 & 54.8
        & 75.5 & 69.1 & 48.0 & 56.7
        & 62.9 & 61.2 \\
        SIDA~\cite{huang2025sidasocialmediaimage}
        & 86.6 & 74.8 & 90.3 & 81.8
        & 81.8 & 74.4 & 69.1 & 71.7
        & 85.1 & 82.0 & 70.7 & 75.9
        & 76.7 & 76.5 \\
        AIGI-Holmes~\cite{zhou2025aigi}
        & 81.0 & 71.2 & 72.4 & 71.8
        & 80.1 & 68.1 & 75.9 & 71.8
        & 83.2 & 78.6 & 68.3 & 73.1
        & 72.2 & 72.1 \\
        LGrad~\cite{DBLP:conf/cvpr/Tan0WGW23}
        & 86.8 & 80.1 & 80.4 & 80.3
        & 86.5 & 80.1 & 79.0 & 79.6
        & 83.5 & 75.0 & 75.7 & 75.3
        & 78.4 & 78.4 \\       
        LNP~\cite{DBLP:journals/corr/abs-2311-00962}
        & 80.1 & 70.8 & 68.4 & 69.6
        & 76.0 & 63.0 & 67.6 & 65.2
        & 82.8 & 75.2 & 72.1 & 73.6
        & 69.4 & 69.5 \\  
        Effort~\cite{yan2024orthogonal}
        & 89.9 & 81.7 & 89.7 & 85.5
        & 86.1 & 78.7 & 79.9 & 79.3
        & 86.4 & 83.4 & 74.0 & 78.4
        & 81.2 & 81.1 \\ 
        DDA~\cite{chen2025dual}
        & 87.0 & 76.4 & 88.1 & 81.8
        & 84.0 & 77.0 & 74.4 & 75.7
        & 85.0 & 81.3 & 71.4 & 76.0
        & 78.0 & 77.8 \\ 
        UnivFD~\cite{DBLP:conf/cvpr/OjhaLL23}
        & \textbf{95.3} & \textbf{90.5} & \textbf{96.1} & \textbf{93.2}
        & 82.1 & 75.8 & 68.3 & 71.8
        & 84.8 & 76.3 & 79.0 & 77.6
        & 81.1 & 80.9 \\ 
        \midrule
        \rowcolor{lightgray!50}
        \textbf{ForenAgent}
        & {94.0} & {87.9} & {95.3} & {91.4}
        & \textbf{93.6} & \textbf{91.3} & \textbf{89.4} & \textbf{90.3}
        & \textbf{91.3} & \textbf{89.5} & \textbf{83.7} & \textbf{86.5}
        & \textbf{89.5} & \textbf{89.4} \\
        \bottomrule
    \end{tabular}
    }
    \label{tabel:detection}
\end{table*}

\section{Experiments}
\subsection{Experimental Setup}
\textbf{Baselines.} 
We compare ForenAgent against 24 baselines across four groups: (1) closed-source MLLMs evaluated zero-shot without fine-tuning, (2) large-scale open-source MLLMs evaluated zero-shot without fine-tuning, (3) tool-augmented MLLMs equipped with our toolbox, and (4) trained baselines, including MLLMs trained on the same image set as ForenAgent and representative state-of-the-art forgery detectors. Detailed baseline descriptions and configurations are provided in the \textit{Supplementary Materials}. Notably, Our evaluation adopts a more challenging three-class protocol; most conventional detectors are binary by design and thus cannot be directly applied without substantial modification.
\\
\textbf{Implementation Details.} 
All experiments are conducted with PyTorch on eight NVIDIA Tesla H200 GPUs; we adopt Qwen2.5-VL-7B as the base MLLM, perform full-parameter finetuning in the Cold-Start stage with a learning rate of 1e-5 for two epochs using AdamW and a cosine-annealing scheduler with a maximum context length of 100k tokens, and then run RFT with GRPO on Qwen2.5-VL-7B for 80 iterations, sampling 256 prompts per batch with eight rollouts per prompt and at most seven tool calls and capping the response length at 20,480 tokens. More details are provided in the \textit{Supplementary Materials}. 
\\
\textbf{Evaluation Metrics.} Following prior work~\cite{huang2025sidasocialmediaimage}, we evaluate detection at the image level using Accuracy and F1. 


\begin{table*}[!t]
\centering
\begin{minipage}[t]{0.48\textwidth}
\centering
\caption{Overall accuracy (\%) and F1-score comparison with state-of-the-art methods on the SIDA-Test dataset.}
\resizebox{\linewidth}{!}{%
\begin{tabular}{lcc}
\hline
\textbf{Methods} & \textbf{Accuracy} & \textbf{F1-score}  \\
\hline
Qwen2.5-VL-7B~\cite{Qwen2.5-VL} & 72.9 & 69.9  \\ 
Qwen3-VL-8B~\cite{Qwen3-VL} & 68.7 & 65.5  \\ 
Gram-Net~\cite{DBLP:conf/cvpr/LiuQT20} & 53.4 & 55.0  \\ 
SIDA~\cite{huang2025sidasocialmediaimage} & 77.2 & 77.1  \\
Effort~\cite{yan2024orthogonal} & 78.5 & 78.4  \\
DDA~\cite{chen2025dual} & 74.3 & 72.9  \\
LGrad~\cite{DBLP:conf/cvpr/Tan0WGW23} & 64.5 & 64.5  \\
LNP~\cite{DBLP:journals/corr/abs-2311-00962} & 53.3 & 53.2   \\ 
UnivFD~\cite{DBLP:conf/cvpr/OjhaLL23} & 61.1 & 60.9   \\ 
\hline
\rowcolor{lightgray!50}
\textbf{ForenAgent} & \textbf{83.1} & \textbf{82.9}  \\         
\hline
\end{tabular}%
}
\label{tab:comparison}
\end{minipage}
\hfill
\begin{minipage}[t]{0.48\textwidth}
\centering
\caption{Evaluation of the influence of different components of ForenAgent on the FABench dataset.}
\resizebox{\linewidth}{!}{%
\begin{tabular}{lcc}
\hline
\textbf{Methods} & \textbf{Accuracy} & \textbf{F1-score}  \\
\hline
w/o Cold Start & 78.7 & 76.9  \\ 
w/o RFT & 81.4 & 81.3  \\ 
w/o Tool Reward & 83.8 & 82.9  \\ 
\hline
w/o global & 86.4 & 86.3  \\ 
w/o logic & 84.4 & 84.3  \\ 
w/o crop & 87.8 & 86.5  \\ 
w/o coh & 88.0 & 87.9  \\ 
\hline
with InternVL3-8B & 79.8 & 79.9  \\ 
with LLaVA-7B & 75.8 & 75.5  \\ 
\hline
\rowcolor{lightgray!50}
\textbf{ForenAgent} & \textbf{89.5} & \textbf{89.4}  \\         
\hline
\end{tabular}%
}
\label{tab:ablation}
\end{minipage}
\end{table*}

\subsection{Detection Evaluation}
As shown in Table~\ref{tabel:detection}, ForenAgent achieves state-of-the-art performance across all test splits, consistently outperforming baselines in both synthetic and tampered categories. 
In contrast, existing closed and open-source MLLMs struggle in zero-shot settings by frequently misclassifying manipulated images as authentic. This highlights a critical deficit of IFD-specific knowledge in their pretraining corpora. 
Notably, equipping models such as Gemini2.5-Pro and GPT-4.1 with specialized forensic tools consistently improves performance, underscoring the potential of tool-augmented IFDL.
After supervised training, Qwen2.5-VL-7B outperforms Qwen3-VL-8B, supporting our choice of Qwen2.5-VL as the backbone. 
UnivFD achieves the best accuracy on the authentic class, suggesting strong capability in identifying unmanipulated images, but it struggles to distinguish between tampered and synthetic cases. 
Overall, these results underline the robustness of ForenAgent across manipulation types and domains, demonstrating its potential as a general-purpose and high-performance solution for real-world IFD.

\subsection{Generalization}
To assess generalization, we further evaluate ForenAgent on the SIDA-Test dataset in Table~\ref{tab:comparison}. Under identical training data, we compare ForenAgent with other methods. ForenAgent achieves the highest scores, demonstrating its strong adaptive capacity. 
The \textit{Supplementary Materials} further explore the limitations of current MLLMs in using bounding boxes for forgery localization and detail the optimal pipeline for adapting ForenAgent to pixel-level precision.

\subsection{Ablation Study}
As shown in Table~\ref{tabel:detection}, ForenAgent substantially outperforms the trained Qwen2.5-VL-7B baseline, confirming that its Python-based low-level toolchain leads to more accurate and robust solutions for IFD. Table \ref{tab:ablation} further presents ablation studies evaluating the contribution of each training stage. Removing the Cold-Start stage (\textit{w/o Cold Start}) noticeably degrades reasoning quality, while removing RFT (\textit{w/o RFT}), which uses our verifiable reward under GRPO, significantly harms final prediction accuracy. 
In addition, we verify the effectiveness of the tool reward. Removing it during RFT (\textit{w/o Tool Reward}) weakens the model’s incentive to utilize tools properly and leads to performance degradation. Overall, these findings highlight the critical role of our staged training and reward design in progressively enhancing reasoning capability.

We also investigate the impact of specific reward components and backbone architectures. Systematic exclusion of the four reward functions: Global Forensic Priority ($R_{\text{global}}$), Tool Logic ($R_{\text{logic}}$), Crop Sensitivity ($R_{\text{crop}}$), and Reasoning Coherence ($R_{\text{coh}}$). Notably, the \textit{w/o logic} variant exhibits the most significant drop in both accuracy and F1-score, suggesting that maintaining syntactically and logically correct tool usage is the foundation of agent efficacy. The degradation in \textit{w/o global} and \textit{w/o coh} further validates our "locate-then-investigate" reasoning chain. 
Moreover, replacing the default architecture with InternVL (\textit{with InternVL3-8B}) or LLaVA (\textit{with LLaVA-7B}) results in substantial performance loss, which confirms that our selected backbone is better suited for the IFD task.
Collectively, these findings underscore the critical synergy between our staged training, structured reward design, and backbone selection.

\begin{figure*}[t]
    \centering
    \includegraphics[width=1\textwidth]{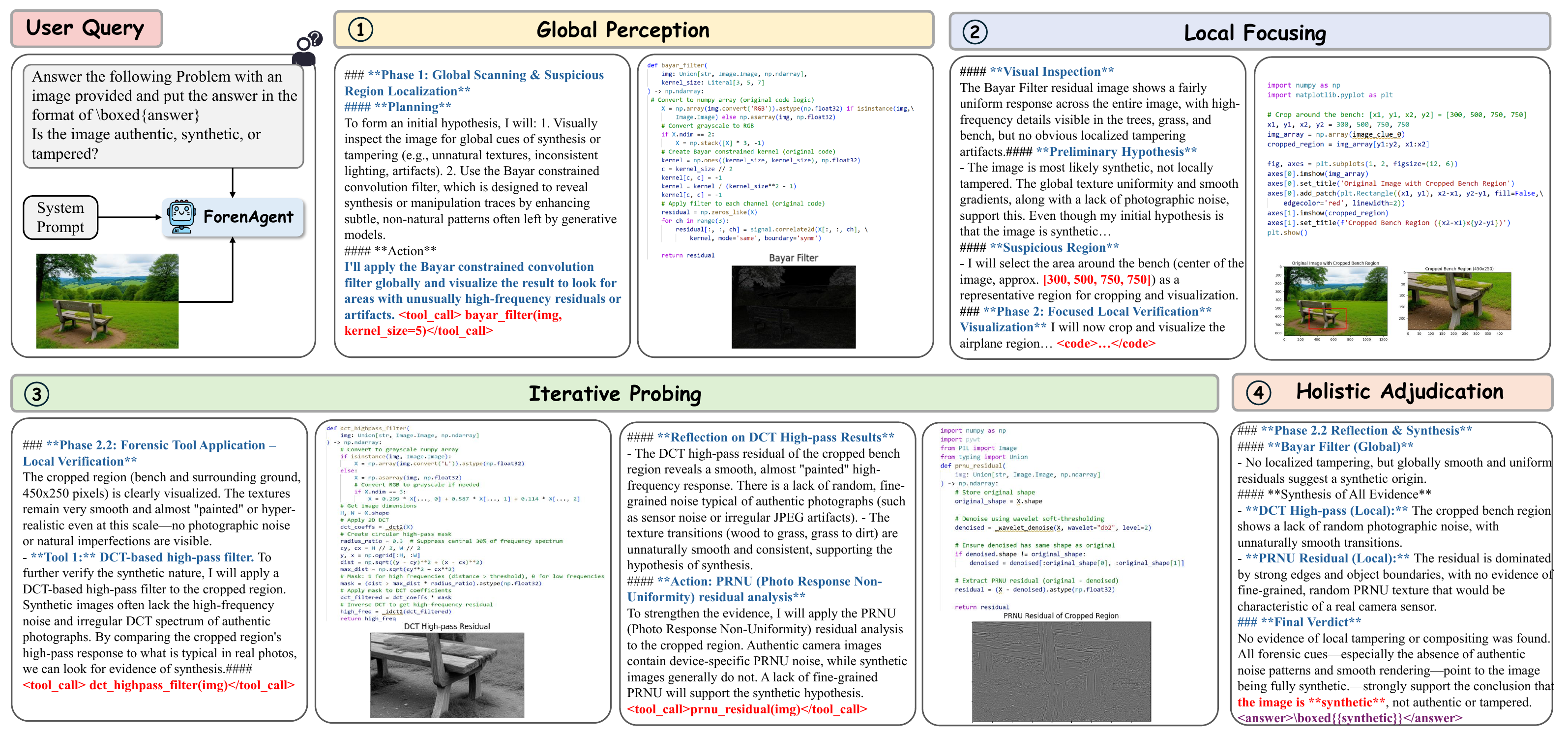}
    \caption{The evidence chain by which ForenAgent correctly identifies a synthetic image.}
    \label{fig:visual}
\end{figure*}

\begin{figure}[!t]
\centering
\begin{minipage}[t]{0.48\linewidth}
    \centering
    \includegraphics[width=\linewidth]{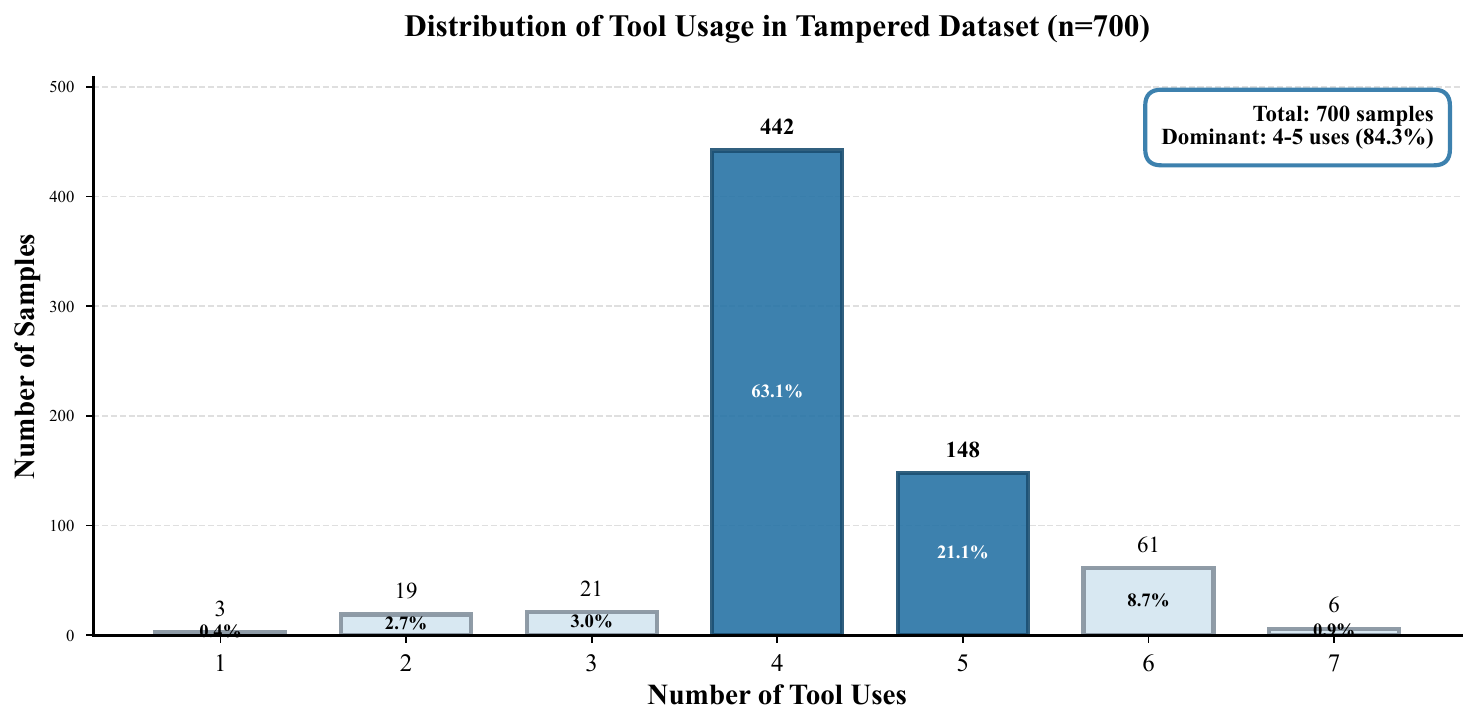}
\end{minipage}
\hfill
\begin{minipage}[t]{0.48\linewidth}
    \centering
    \includegraphics[width=\linewidth]{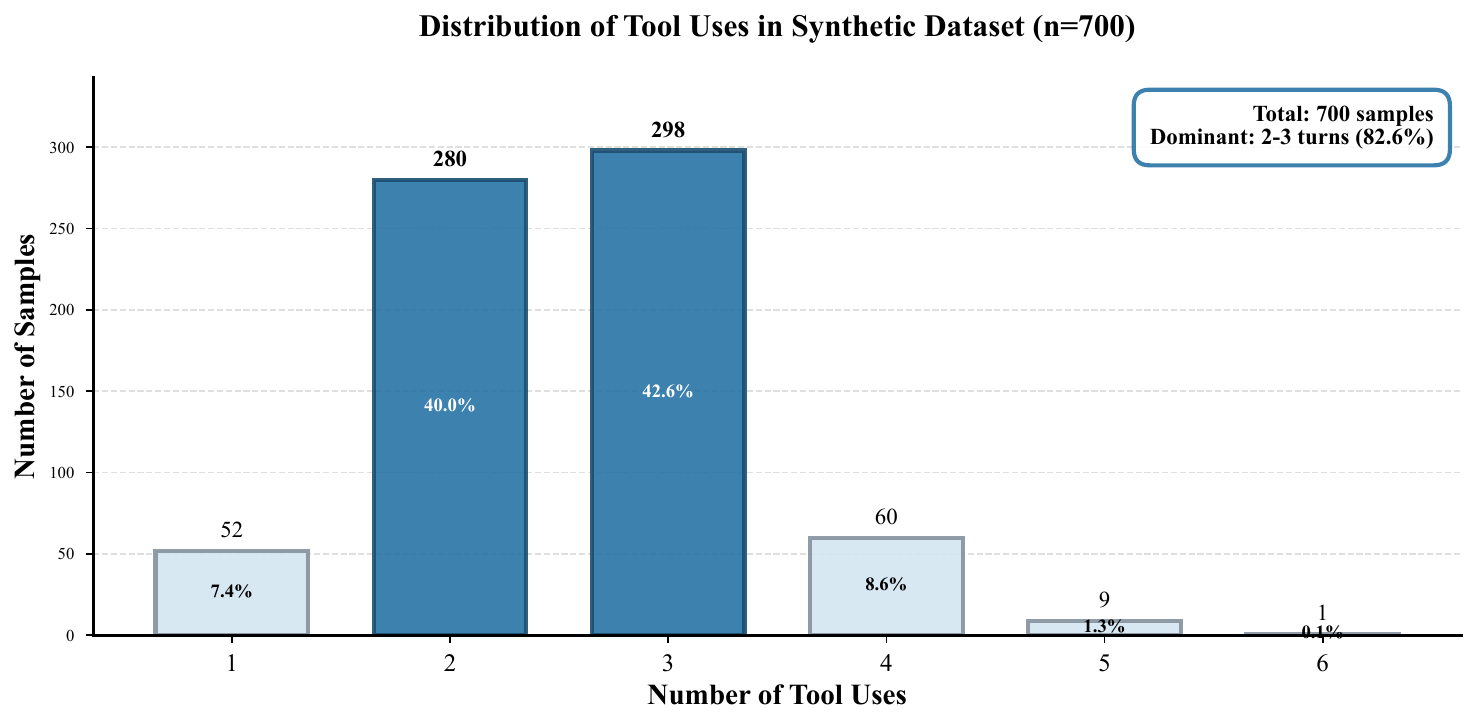}
\end{minipage}
\caption{Distribution of tool invocation frequencies for ForenAgent across FABench.}
\label{fig:distribution_analysis}
\end{figure}

\subsection{Visualization}
Figure~\ref{fig:visual} illustrates a successful synthetic-image case where ForenAgent constructs a coherent evidence chain mirroring human reasoning. The process progresses through global perception, local focusing, iterative probing, and holistic adjudication to deliver accurate detection and a convincing explanation. It first applies the Bayar filter for global scanning to identify initial suspicion based on global texture uniformity and lack of photographic noise. The agent then conducts local focusing on a representative region and iteratively probes with DCT high-pass filtering and PRNU residual analysis. Finally, the agent synthesizes multi-scale forensic cues, concluding that the hyper-realistic smooth rendering and lack of random noise point to a fully synthetic origin. 

\subsection{Tool Usage Analysis}

\begin{wrapfigure}{r}{0.6\textwidth}
    \centering
    \vspace{-5mm}
    \includegraphics[width=0.48\textwidth]{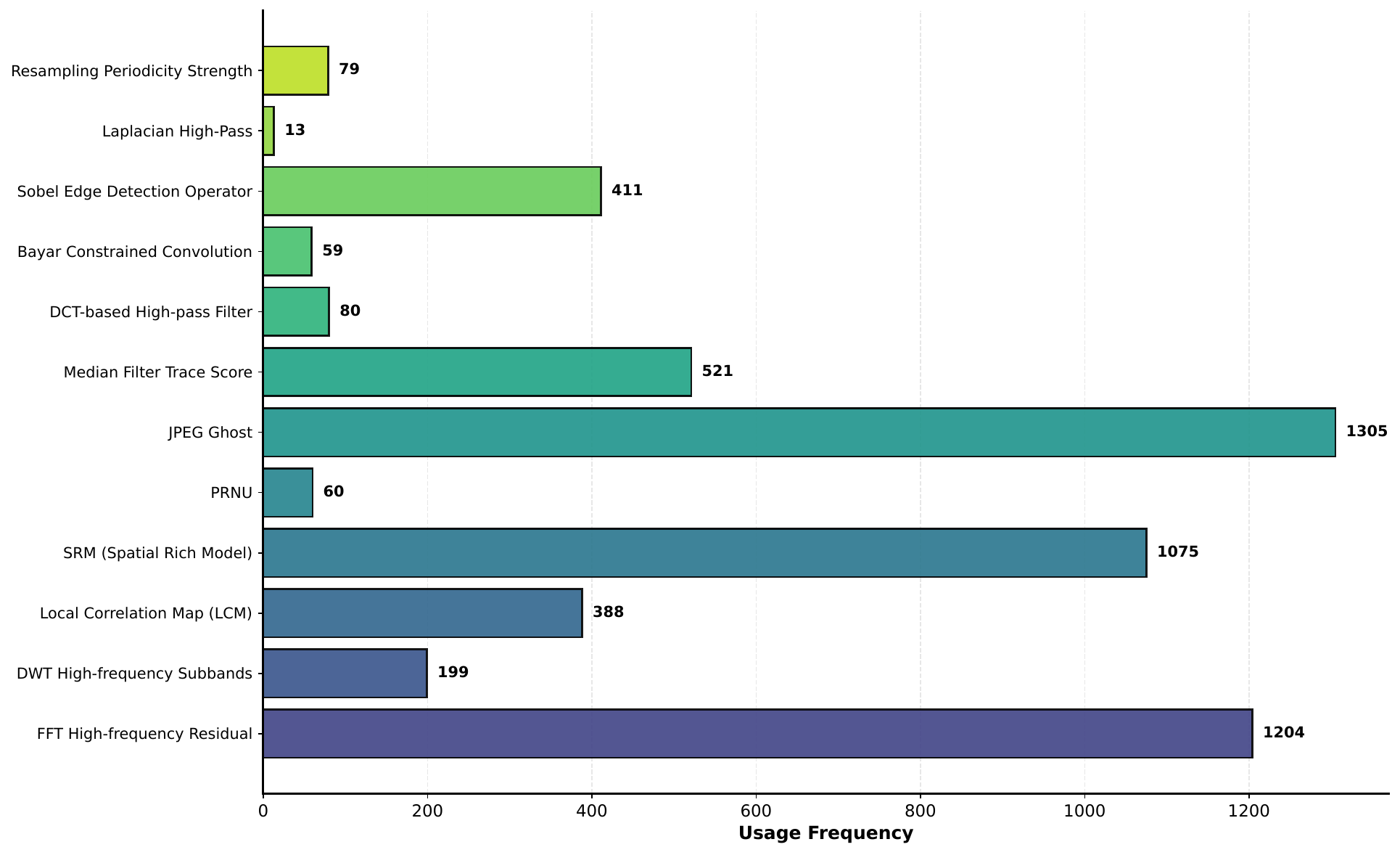}
    \caption{Quantitative distribution of low-level forensic tool usage frequencies across the comprehensive FABench dataset.}
    \label{fig:tool_usage_frequency_final}
    \vspace{-5mm}
\end{wrapfigure}

Figure~\ref{fig:distribution_analysis} analyzes tool usage frequencies on the FABench test sets. For synthetic images, ForenAgent typically converges within about $3$ tool calls, whereas tampered images require around $4$ calls due to higher structural complexity and the need to localize manipulated regions. This suggests that ForenAgent dynamically adapts its reasoning workflow to task difficulty, substantially improving efficiency.
Figure~\ref{fig:tool_usage_frequency_final} summarizes the usage distribution across low-level forensic tools: SRM, FFT, and JPEG Ghost are used most frequently, while Laplacian High-Pass appears less often. This pattern suggests that ForenAgent learns an adaptive tool-selection policy conditioned on image characteristics, rather than relying on mechanical tool enumeration. 
Notably, many classical low-level forensic tools are revived and integrated into our pipeline, offering a new perspective on combining traditional image forgery detection techniques with modern MLLMs and potentially inspiring future hybrid-agent designs.
Detailed evaluations of computational efficiency and robustness are discussed in the \textit{Supplementary Materials}.

\section{Conclusion}
In this paper, we introduced ForenAgent, an interactive multi-round framework that enables MLLMs to autonomously invoke and iteratively refine Python-based low-level tools for IFD. 
We abstract and generalize key low-level tools in IFD, forming a 12-tool forensic toolbox for future community extension.
Through a two-stage training pipeline of Cold Start and Reinforcement Fine-Tuning, ForenAgent learns a dynamic reasoning process from global perception to holistic adjudication. Experimental results on FABench and SIDA-Test demonstrate its superior interpretability, robustness, and reflective tool-use capability across diverse forgery scenarios.
Our work marks an important first step toward building intelligent agent systems for image forensics.


\section*{Acknowledgments}
This work was supported by the National Natural Science Foundation
of China (NSFC) under Grant 62476260, the Fundamental Research Funds for the Central Universities under Grant WK2100000057.



%
%
\bibliographystyle{splncs04}
\bibliography{main}

@String(CVPR= {IEEE Conf. Comput. Vis. Pattern Recog.})

@String(ICLR = {Int. Conf. Learn. Represent.})

@String(AAAI = {AAAI})

@String(CVPR  = {CVPR})

@String(ICLR  = {ICLR})

@inproceedings{DBLP:conf/cvpr/Tan0WGW23,
  author       = {Chuangchuang Tan and
                  Yao Zhao and
                  Shikui Wei and
                  Guanghua Gu and
                  Yunchao Wei},
  title        = {Learning on Gradients: Generalized Artifacts Representation for GAN-Generated Images Detection},
  booktitle    = CVPR,
  year         = {2023}
}

@inproceedings{DBLP:conf/cvpr/OjhaLL23,
  author       = {Utkarsh Ojha and
                  Yuheng Li and
                  Yong Jae Lee},
  title        = {Towards Universal Fake Image Detectors that Generalize Across Generative
                  Models},
  booktitle    = CVPR,
  year         = {2023}
}

@inproceedings{DBLP:conf/cvpr/LiuQT20,
  author       = {Zhengzhe Liu and
                  Xiaojuan Qi and
                  Philip H. S. Torr},
  title        = {Global Texture Enhancement for Fake Face Detection in the Wild},
  booktitle    = {CVPR},
  year         = {2020}
}

@article{team2023gemini,
  title={Gemini: a family of highly capable multimodal models},
  author={Team, Gemini and Anil, Rohan and Borgeaud, Sebastian and Alayrac, Jean-Baptiste and Yu, Jiahui and Soricut, Radu and Schalkwyk, Johan and Dai, Andrew M and Hauth, Anja and Millican, Katie and others},
  journal={arXiv preprint arXiv:2312.11805},
  year={2023}
}

@inproceedings{xu2024fakeshield,
        title={FakeShield: Explainable Image Forgery Detection and Localization via Multi-modal Large Language Models},
        author={Xu, Zhipei and Zhang, Xuanyu and Li, Runyi and Tang, Zecheng and Huang, Qing and Zhang, Jian},
        booktitle={International Conference on Learning Representations},
        year={2025}
}

@inproceedings{luo2021generalizing,
  title={Generalizing face forgery detection with high-frequency features},
  author={Luo, Yuchen and Zhang, Yong and Yan, Junchi and Liu, Wei},
  booktitle={Proceedings of the IEEE/CVF conference on computer vision and pattern recognition},
  pages={16317--16326},
  year={2021}
}

@article{zhang2023prnu,
  title={PRNU-based image forgery localization with deep multi-scale fusion},
  author={Zhang, Yushu and Tan, Qing and Qi, Shuren and Xue, Mingfu},
  journal={ACM Transactions on Multimedia Computing, Communications and Applications},
  volume={19},
  number={2},
  pages={1--20},
  year={2023},
  publisher={ACM New York, NY}
}

@article{li2025fakescope,
  title={Fakescope: Large multimodal expert model for transparent ai-generated image forensics},
  author={Li, Yixuan and Tian, Yu and Huang, Yipo and Lu, Wei and Wang, Shiqi and Lin, Weisi and Rocha, Anderson},
  journal={arXiv preprint arXiv:2503.24267},
  year={2025}
}

@article{kang2025legion,
  title={Legion: Learning to ground and explain for synthetic image detection},
  author={Kang, Hengrui and Wen, Siwei and Wen, Zichen and Ye, Junyan and Li, Weijia and Feng, Peilin and Zhou, Baichuan and Wang, Bin and Lin, Dahua and Zhang, Linfeng and others},
  journal={arXiv preprint arXiv:2503.15264},
  year={2025}
}

@article{liu2024forgerygpt,
  title={Forgerygpt: Multimodal large language model for explainable image forgery detection and localization},
  author={Liu, Jiawei and Zhang, Fanrui and Zhu, Jiaying and Sun, Esther and Zhang, Qiang and Zha, Zheng-Jun},
  journal={arXiv preprint arXiv:2410.10238},
  year={2024}
}

@article{popescu2005exposing,
  title={Exposing digital forgeries by detecting traces of resampling},
  author={Popescu, Alin C and Farid, Hany},
  journal={IEEE Transactions on signal processing},
  volume={53},
  number={2},
  pages={758--767},
  year={2005},
  publisher={IEEE}
}

@inproceedings{tan2024frequency,
  title={Frequency-aware deepfake detection: Improving generalizability through frequency space domain learning},
  author={Tan, Chuangchuang and Zhao, Yao and Wei, Shikui and Gu, Guanghua and Liu, Ping and Wei, Yunchao},
  booktitle={Proceedings of the AAAI Conference on Artificial Intelligence},
  volume={38},
  number={5},
  pages={5052--5060},
  year={2024}
}

@inproceedings{li2025improving,
  title={Improving synthetic image detection towards generalization: An image transformation perspective},
  author={Li, Ouxiang and Cai, Jiayin and Hao, Yanbin and Jiang, Xiaolong and Hu, Yao and Feng, Fuli},
  booktitle={Proceedings of the 31st ACM SIGKDD Conference on Knowledge Discovery and Data Mining V. 1},
  pages={2405--2414},
  year={2025}
}

@inproceedings{huang2025sidasocialmediaimage,
  title={Sida: Social media image deepfake detection, localization and explanation with large multimodal model},
  author={Huang, Zhenglin and Hu, Jinwei and Li, Xiangtai and He, Yiwei and Zhao, Xingyu and Peng, Bei and Wu, Baoyuan and Huang, Xiaowei and Cheng, Guangliang},
  booktitle={Proceedings of the Computer Vision and Pattern Recognition Conference},
  pages={28831--28841},
  year={2025}
}

@article{achiam2023gpt,
  title={Gpt-4 technical report},
  author={Achiam, Josh and Adler, Steven and Agarwal, Sandhini and Ahmad, Lama and Akkaya, Ilge and Aleman, Florencia Leoni and Almeida, Diogo and Altenschmidt, Janko and Altman, Sam and Anadkat, Shyamal and others},
  journal={arXiv preprint arXiv:2303.08774},
  year={2023}
}

@article{jaech2024openai,
  title={Openai o1 system card},
  author={Jaech, Aaron and Kalai, Adam and Lerer, Adam and Richardson, Adam and El-Kishky, Ahmed and Low, Aiden and Helyar, Alec and Madry, Aleksander and Beutel, Alex and Carney, Alex and others},
  journal={arXiv preprint arXiv:2412.16720},
  year={2024}
}

@article{wang2024mpo,
  title={Enhancing the Reasoning Ability of Multimodal Large Language Models via Mixed Preference Optimization},
  author={Wang, Weiyun and Chen, Zhe and Wang, Wenhai and Cao, Yue and Liu, Yangzhou and Gao, Zhangwei and Zhu, Jinguo and Zhu, Xizhou and Lu, Lewei and Qiao, Yu and Dai, Jifeng},
  journal={arXiv preprint arXiv:2411.10442},
  year={2024}
}

@article{wu2025qwenimage,
  title={Qwen-image technical report},
  author={Wu, Chenfei and Li, Jiahao and Zhou, Jingren and Lin, Junyang and Gao, Kaiyuan and Yan, Kun and Yin, Sheng-ming and Bai, Shuai and Xu, Xiao and Chen, Yilei and others},
  journal={arXiv preprint arXiv:2508.02324},
  year={2025}
}

@misc{qvq-72b-preview,
    title = {QVQ: To See the World with Wisdom},
    url = {https://qwenlm.github.io/blog/qvq-72b-preview/},
    author = {Qwen Team},
    month = {December},
    year = {2024},
    note = {Accessed: 2026-06-29}
}

@inproceedings{sdxl,
  author       = {Dustin Podell and
                  Zion English and
                  Kyle Lacey and
                  Andreas Blattmann and
                  Tim Dockhorn and
                  Jonas M{\"{u}}ller and
                  Joe Penna and
                  Robin Rombach},
  title        = {{SDXL:} Improving Latent Diffusion Models for High-Resolution Image
                  Synthesis},
  booktitle    = {ICLR},
  publisher    = {OpenReview.net},
  year         = {2024},
}

@article{guo2025deepseek,
  title={Deepseek-r1: Incentivizing reasoning capability in llms via reinforcement learning},
  author={Guo, Daya and Yang, Dejian and Zhang, Haowei and Song, Junxiao and Zhang, Ruoyu and Xu, Runxin and Zhu, Qihao and Ma, Shirong and Wang, Peiyi and Bi, Xiao and others},
  journal={arXiv preprint arXiv:2501.12948},
  year={2025}
}

@article{Qwen2.5-VL,
  title={Qwen2.5-VL Technical Report},
  author={Bai, Shuai and Chen, Keqin and Liu, Xuejing and Wang, Jialin and Ge, Wenbin and Song, Sibo and Dang, Kai and Wang, Peng and Wang, Shijie and Tang, Jun and Zhong, Humen and Zhu, Yuanzhi and Yang, Mingkun and Li, Zhaohai and Wan, Jianqiang and Wang, Pengfei and Ding, Wei and Fu, Zheren and Xu, Yiheng and Ye, Jiabo and Zhang, Xi and Xie, Tianbao and Cheng, Zesen and Zhang, Hang and Yang, Zhibo and Xu, Haiyang and Lin, Junyang},
  journal={arXiv preprint arXiv:2502.13923},
  year={2025}
}

@inproceedings{gupta2023visual,
  title={Visual programming: Compositional visual reasoning without training},
  author={Gupta, Tanmay and Kembhavi, Aniruddha},
  booktitle={Proceedings of the IEEE/CVF conference on computer vision and pattern recognition},
  pages={14953--14962},
  year={2023}
}

@inproceedings{suris2023vipergpt,
  title={Vipergpt: Visual inference via python execution for reasoning},
  author={Sur{\'\i}s, D{\'\i}dac and Menon, Sachit and Vondrick, Carl},
  booktitle={Proceedings of the IEEE/CVF international conference on computer vision},
  pages={11888--11898},
  year={2023}
}

@article{wang2024metatool,
  title={MetaTool: Facilitating Large Language Models to Master Tools with Meta-task Augmentation},
  author={Wang, Xiaohan and Li, Dian and Zhao, Yilin and Wang, Hui and others},
  journal={arXiv preprint arXiv:2407.12871},
  year={2024}
}

@article{zhao2025pyvision,
  title={Pyvision: Agentic vision with dynamic tooling},
  author={Zhao, Shitian and Zhang, Haoquan and Lin, Shaoheng and Li, Ming and Wu, Qilong and Zhang, Kaipeng and Wei, Chen},
  journal={arXiv preprint arXiv:2507.07998},
  year={2025}
}

@article{zheng2025deepeyes,
  title={DeepEyes: Incentivizing" Thinking with Images" via Reinforcement Learning},
  author={Zheng, Ziwei and Yang, Michael and Hong, Jack and Zhao, Chenxiao and Xu, Guohai and Yang, Le and Shen, Chao and Yu, Xing},
  journal={arXiv preprint arXiv:2505.14362},
  year={2025}
}

@article{su2025openthinkimg,
  title={Openthinkimg: Learning to think with images via visual tool reinforcement learning},
  author={Su, Zhaochen and Li, Linjie and Song, Mingyang and Hao, Yunzhuo and Yang, Zhengyuan and Zhang, Jun and Chen, Guanjie and Gu, Jiawei and Li, Juntao and Qu, Xiaoye and others},
  journal={arXiv preprint arXiv:2505.08617},
  year={2025}
}

@article{zhou2025reinforced,
  title={Reinforced visual perception with tools},
  author={Zhou, Zetong and Chen, Dongping and Ma, Zixian and Hu, Zhihan and Fu, Mingyang and Wang, Sinan and Wan, Yao and Zhao, Zhou and Krishna, Ranjay},
  journal={arXiv preprint arXiv:2509.01656},
  year={2025}
}

@article{liu2025visual,
  title={Visual Agentic Reinforcement Fine-Tuning},
  author={Liu, Ziyu and Zang, Yuhang and Zou, Yushan and Liang, Zijian and Dong, Xiaoyi and Cao, Yuhang and Duan, Haodong and Lin, Dahua and Wang, Jiaqi},
  journal={arXiv preprint arXiv:2505.14246},
  year={2025}
}

@article{zhu2024intelligent,
  title={An intelligent agentic system for complex image restoration problems},
  author={Zhu, Kaiwen and Gu, Jinjin and You, Zhiyuan and Qiao, Yu and Dong, Chao},
  journal={arXiv preprint arXiv:2410.17809},
  year={2024}
}

@inproceedings{MVSS-Net2021image,
  title={Image manipulation detection by multi-view multi-scale supervision},
  author={Chen, Xinru and Dong, Chengbo and Ji, Jiaqi and Cao, Juan and Li, Xirong},
  booktitle={Proceedings of the IEEE/CVF International Conference on Computer Vision},
  pages={14185--14193},
  year={2021}
}

@inproceedings{Objectformerwang2022objectformer,
  title={Objectformer for image manipulation detection and localization},
  author={Wang, Junke and Wu, Zuxuan and Chen, Jingjing and Han, Xintong and Shrivastava, Abhinav and Lim, Ser-Nam and Jiang, Yu-Gang},
  booktitle={Proceedings of the IEEE/CVF Conference on Computer Vision and Pattern Recognition},
  pages={2364--2373},
  year={2022}
}

@InProceedings{HiFi_IFDL,
    author    = {Guo, Xiao and Liu, Xiaohong and Ren, Zhiyuan and Grosz, Steven and Masi, Iacopo and Liu, Xiaoming},
    title     = {Hierarchical Fine-Grained Image Forgery Detection and Localization},
    booktitle = {Proceedings of the IEEE/CVF Conference on Computer Vision and Pattern Recognition},
    month     = {June},
    year      = {2023},
    pages     = {3155-3165}
}

@article{zhou2025aigi,
  title={AIGI-Holmes: Towards Explainable and Generalizable AI-Generated Image Detection via Multimodal Large Language Models},
  author={Zhou, Ziyin and Luo, Yunpeng and Wu, Yuanchen and Sun, Ke and Ji, Jiayi and Yan, Ke and Ding, Shouhong and Sun, Xiaoshuai and Wu, Yunsheng and Ji, Rongrong},
  journal={arXiv preprint arXiv:2507.02664},
  year={2025}
}

@article{ye2024loki,
  title={Loki: A comprehensive synthetic data detection benchmark using large multimodal models},
  author={Ye, Junyan and Zhou, Baichuan and Huang, Zilong and Zhang, Junan and Bai, Tianyi and Kang, Hengrui and He, Jun and Lin, Honglin and Wang, Zihao and Wu, Tong and others},
  journal={arXiv preprint arXiv:2410.09732},
  year={2024}
}

@article{yan2024orthogonal,
  title={Orthogonal Subspace Decomposition for Generalizable AI-Generated Image Detection},
  author={Yan, Zhiyuan and Wang, Jiangming and Jin, Peng and Zhang, Ke-Yue and Liu, Chengchun and Chen, Shen and Yao, Taiping and Ding, Shouhong and Wu, Baoyuan and Yuan, Li},
  journal={arXiv preprint arXiv:2411.15633},
  year={2024}
}

@article{Qwen3-VL,
      title={Qwen3-VL Technical Report}, 
      author={Shuai Bai and Yuxuan Cai and Ruizhe Chen and Keqin Chen and Xionghui Chen and Zesen Cheng and Lianghao Deng and Wei Ding and Chang Gao and Chunjiang Ge and Wenbin Ge and Zhifang Guo and Qidong Huang and Jie Huang and Fei Huang and Binyuan Hui and Shutong Jiang and Zhaohai Li and Mingsheng Li and Mei Li and Kaixin Li and Zicheng Lin and Junyang Lin and Xuejing Liu and Jiawei Liu and Chenglong Liu and Yang Liu and Dayiheng Liu and Shixuan Liu and Dunjie Lu and Ruilin Luo and Chenxu Lv and Rui Men and Lingchen Meng and Xuancheng Ren and Xingzhang Ren and Sibo Song and Yuchong Sun and Jun Tang and Jianhong Tu and Jianqiang Wan and Peng Wang and Pengfei Wang and Qiuyue Wang and Yuxuan Wang and Tianbao Xie and Yiheng Xu and Haiyang Xu and Jin Xu and Zhibo Yang and Mingkun Yang and Jianxin Yang and An Yang and Bowen Yu and Fei Zhang and Hang Zhang and Xi Zhang and Bo Zheng and Humen Zhong and Jingren Zhou and Fan Zhou and Jing Zhou and Yuanzhi Zhu and Ke Zhu},
	  journal={arXiv preprint arXiv:2511.21631},
      year={2025}
}

@article{chen2025dual,
  title={Dual Data Alignment Makes AI-Generated Image Detector Easier Generalizable},
  author={Chen, Ruoxin and Xi, Junwei and Yan, Zhiyuan and Zhang, Ke-Yue and Wu, Shuang and Xie, Jingyi and Chen, Xu and Xu, Lei and Guan, Isabel and Yao, Taiping and others},
  journal={arXiv preprint arXiv:2505.14359},
  year={2025}
}

@article{zhang2025thyme,
  title={Thyme: Think beyond images},
  author={Zhang, Yi-Fan and Lu, Xingyu and Yin, Shukang and Fu, Chaoyou and Chen, Wei and Hu, Xiao and Wen, Bin and Jiang, Kaiyu and Liu, Changyi and Zhang, Tianke and others},
  journal={arXiv preprint arXiv:2508.11630},
  year={2025}
}

@article{qiao2024fully,
  title={Fully unsupervised deepfake video detection via enhanced contrastive learning},
  author={Qiao, Tong and Xie, Shichuang and Chen, Yanli and Retraint, Florent and Luo, Xiangyang},
  journal={IEEE Transactions on Pattern Analysis and Machine Intelligence},
  year={2024},
  publisher={IEEE}
}

@article{survey-mllm,
  author       = {Shukang Yin and
                  Chaoyou Fu and
                  Sirui Zhao and
                  Ke Li and
                  Xing Sun and
                  Tong Xu and
                  Enhong Chen},
  title        = {A Survey on Multimodal Large Language Models},
  journal      = {CoRR},
  volume       = {abs/2306.13549},
  year         = {2023},
  url          = {https://doi.org/10.48550/arXiv.2306.13549},
  doi          = {10.48550/ARXIV.2306.13549},
  eprinttype    = {arXiv},
  eprint       = {2306.13549},
  bibsource    = {dblp computer science bibliography, https://dblp.org}
}

@article{rao2022towards,
  title={Towards JPEG-resistant image forgery detection and localization via self-supervised domain adaptation},
  author={Rao, Yuan and Ni, Jiangqun and Zhang, Weizhe and Huang, Jiwu},
  journal={IEEE transactions on pattern analysis and machine intelligence},
  volume={47},
  number={5},
  pages={3285--3297},
  year={2022},
  publisher={IEEE}
}

@inproceedings{karras2019style,
  title={A style-based generator architecture for generative adversarial networks},
  author={Karras, Tero and Laine, Samuli and Aila, Timo},
  booktitle={Proceedings of the IEEE/CVF conference on computer vision and pattern recognition},
  pages={4401--4410},
  year={2019}
}

@article{shao2024detecting,
  title={Detecting and grounding multi-modal media manipulation and beyond},
  author={Shao, Rui and Wu, Tianxing and Wu, Jianlong and Nie, Liqiang and Liu, Ziwei},
  journal={IEEE Transactions on Pattern Analysis and Machine Intelligence},
  year={2024},
  publisher={IEEE}
}

@article{zhu2023face,
  title={Face forgery detection by 3d decomposition and composition search},
  author={Zhu, Xiangyu and Fei, Hongyan and Zhang, Bin and Zhang, Tianshuo and Zhang, Xiaoyu and Li, Stan Z and Lei, Zhen},
  journal={IEEE Transactions on Pattern Analysis and Machine Intelligence},
  volume={45},
  number={7},
  pages={8342--8357},
  year={2023},
  publisher={IEEE}
}

@article{qi2022principled,
  title={A principled design of image representation: Towards forensic tasks},
  author={Qi, Shuren and Zhang, Yushu and Wang, Chao and Zhou, Jiantao and Cao, Xiaochun},
  journal={IEEE Transactions on Pattern Analysis and Machine Intelligence},
  volume={45},
  number={5},
  pages={5337--5354},
  year={2022},
  publisher={IEEE}
}

@inproceedings{zhang2024common,
  title={Common sense reasoning for deepfake detection},
  author={Zhang, Yue and Colman, Ben and Guo, Xiao and Shahriyari, Ali and Bharaj, Gaurav},
  booktitle={European Conference on Computer Vision},
  pages={399--415},
  year={2024},
  organization={Springer}
}

@article{huang2025so,
  title={So-Fake: Benchmarking and Explaining Social Media Image Forgery Detection},
  author={Huang, Zhenglin and Li, Tianxiao and Li, Xiangtai and Wen, Haiquan and He, Yiwei and Zhang, Jiangning and Fei, Hao and Yang, Xi and Huang, Xiaowei and Peng, Bei and others},
  journal={arXiv preprint arXiv:2505.18660},
  year={2025}
}

@inproceedings{lin2014microsoft,
  title={Microsoft coco: Common objects in context},
  author={Lin, Tsung-Yi and Maire, Michael and Belongie, Serge and Hays, James and Perona, Pietro and Ramanan, Deva and Doll{\'a}r, Piotr and Zitnick, C Lawrence},
  booktitle={European conference on computer vision},
  pages={740--755},
  year={2014},
  organization={Springer}
}

@inproceedings{chen2022self,
  title={Self-supervised learning of adversarial example: Towards good generalizations for deepfake detection},
  author={Chen, Liang and Zhang, Yong and Song, Yibing and Liu, Lingqiao and Wang, Jue},
  booktitle={Proceedings of the IEEE/CVF conference on computer vision and pattern recognition},
  pages={18710--18719},
  year={2022}
}

@article{DBLP:journals/corr/abs-2311-00962,
  title={Detecting generated images by real images only},
  author={Bi, Xiuli and Liu, Bo and Yang, Fan and Xiao, Bin and Li, Weisheng and Huang, Gao and Cosman, Pamela C},
  journal={arXiv preprint arXiv:2311.00962},
  year={2023}
}

@misc{openai2025gpt41,
  author       = {OpenAI},
  title        = {Introducing GPT-4.1},
  year         = {2025},
  howpublished = {\url{https://openai.com/index/gpt-4-1/}},
  note         = {Accessed: 2025-11-13}
}

@misc{openai2025gpt5,
  title        = {GPT-5 and the New Era of Work},
  author       = {{OpenAI}},
  year         = {2025},
  month        = {aug},
  howpublished = {\url{https://openai.com/index/gpt-5-new-era-of-work/}},
  note         = {Accessed: 2026-02-28}
}

@article{zhu2025internvl3,
  title={Internvl3: Exploring advanced training and test-time recipes for open-source multimodal models},
  author={Zhu, Jinguo and Wang, Weiyun and Chen, Zhe and Liu, Zhaoyang and Ye, Shenglong and Gu, Lixin and Tian, Hao and Duan, Yuchen and Su, Weijie and Shao, Jie and others},
  journal={arXiv preprint arXiv:2504.10479},
  year={2025}
}

@misc{flux2024,
  author = {Black Forest Labs},
  title = {FLUX},
  year = {2024},
  howpublished = {\url{https://github.com/black-forest-labs/flux}},
  note = {Accessed: 2026-06-29}
}

@misc{midjourney2023,
  author = {{Midjourney}},
  title = {Midjourney (Version 7)},
  year = {2025},
  howpublished = {\url{https://www.midjourney.com/}},
  note = {Accessed: 2026-06-29}
}

@misc{nano,
  author       = {{Google}},
  title        = {{Nanobanana}},
  howpublished = {\url{https://developers.googleblog.com/en/introducing-gemini-2-5-flash-image/}},
  year         = {2025},
  note         = {Accessed: 2026-06-29}
}

@misc{openai2024dalle3,
  author       = {{OpenAI}},
  title        = {{DALL{\textperiodcentered}E 3}},
  howpublished = {\url{https://openai.com/dall-e}},
  year         = {2024},
  note         = {Accessed: 2026-06-29}
}

@inproceedings{liu2024forgery,
  title={Forgery-aware adaptive transformer for generalizable synthetic image detection},
  author={Liu, Huan and Tan, Zichang and Tan, Chuangchuang and Wei, Yunchao and Wang, Jingdong and Zhao, Yao},
  booktitle={Proceedings of the IEEE/CVF Conference on Computer Vision and Pattern Recognition},
  pages={10770--10780},
  year={2024}
}

@inproceedings{bayar2016deep,
  title={A deep learning approach to universal image manipulation detection using a new convolutional layer},
  author={Bayar, Belhassen and Stamm, Matthew C},
  booktitle={Proceedings of the 4th ACM workshop on information hiding and multimedia security},
  pages={5--10},
  year={2016}
}

@article{hong2025deepeyesv2,
  title={DeepEyesV2: Toward Agentic Multimodal Model},
  author={Hong, Jack and Zhao, Chenxiao and Zhu, ChengLin and Lu, Weiheng and Xu, Guohai and Yu, Xing},
  journal={arXiv preprint arXiv:2511.05271},
  year={2025}
}

@article{yan2024sanity,
  title={A sanity check for ai-generated image detection},
  author={Yan, Shilin and Li, Ouxiang and Cai, Jiayin and Hao, Yanbin and Jiang, Xiaolong and Hu, Yao and Xie, Weidi},
  journal={arXiv preprint arXiv:2406.19435},
  year={2024}
}

@inproceedings{wang2025spatial,
  title={Spatial-Temporal Forgery Trace based Forgery Image Identification},
  author={Wang, Yilin and Feng, Zunlei and Wang, Jiachi and Lou, Hengrui and Zhou, Binjia and Lei, Jie and Song, Mingli and Bei, Yijun},
  booktitle={Proceedings of the IEEE/CVF International Conference on Computer Vision},
  pages={17067--17076},
  year={2025}
}

@inproceedings{FFT,
  title={Thinking in frequency: Face forgery detection by mining frequency-aware clues},
  author={Qian, Yuyang and Yin, Guojun and Sheng, Lu and Chen, Zixuan and Shao, Jing},
  booktitle={European conference on computer vision},
  pages={86--103},
  year={2020},
  organization={Springer}
}

@misc{openai2025o3mini,
  author       = {OpenAI},
  title        = {{OpenAI} o3-mini System Card},
  year         = {2025},
  note         = {Published January 31, 2025. Accessed: 2026-06-29},
  url          = {https://cdn.openai.com/o3-mini-system-card-feb10.pdf}
}
\end{document}